\documentclass{scrartcl} \usepackage[hyphens]{url}  \usepackage{graphicx} \usepackage{lmodern}
\usepackage{hyperref}
\usepackage[format=plain, labelfont=bf]{caption}

\usepackage[backend=bibtex, citestyle=authoryear-comp]{biblatex}
\usepackage{xifthen}

\title{Graph Neural Networks for Maximum Constraint Satisfaction}

\author{Jan Toenshoff \and Martin Ritzert \and Hinrikus Wolf \and Martin Grohe}

\publishers{RWTH Aachen University}

\usepackage{float}
\usepackage{tikz} \usepackage{tikz-network}
\usepackage{microtype}
\usepackage{amsmath,amssymb,amsfonts}
\usepackage[ruled,noend]{algorithm2e}
\usepackage{tikz}
\usepackage{booktabs}
\usepackage{bm}
\usepackage{mathtools}
\usepackage[textwidth=2cm,textsize=small,disable]{todonotes}

\usepackage[noend]{algorithmic}
\renewcommand{\algorithmiccomment}[1]{// \textit{#1}}

\newcommand{\cut}[1]{}
\newcommand{\ie}{\mbox{i.e.}}

\renewcommand{\phi}{\varphi}
\renewcommand{\theta}{\vartheta}
\newcommand{\NN}{\mathbb{N}}
\newcommand{\RR}{\mathbb{R}}
\newcommand{\citeinline}[1]{\citeauthor{#1}~(\citeyear{#1})}
\newcommand{\citet}[1]{\citeinline{#1}}
\renewcommand{\cite}[1]{\parencite{#1}}
\newcommand{\eval}{t_\text{max}^\text{ev}}
\newcommand{\train}{t_\text{max}^\text{tr}}
\newcommand{\tmax}{t_\text{max}}

\DeclareMathOperator{\softmax}{softmax}

\newcommand{\mcol}{\textsc{Max\hbox{-}3\hbox{-}Col}}
\newcommand{\dcol}{\textsc{3\hbox{-}Col}}
\newcommand{\msat}{\textsc{Max\hbox{-}2\hbox{-}Sat}}
\newcommand{\mis}{\textsc{Max\hbox{-}IS}}
\newcommand{\csp}{\textsc{CSP}}
\newcommand{\sat}{\textsc{SAT}}
\newcommand{\tsp}{\textsc{TSP}}
\newcommand{\mcut}{\textsc{Max-Cut}}
\newcommand{\mcsp}[1]{\ifthenelse{\isempty{#1}}{{\textsc{Max\hbox{-}CSP}}}{binary {\textsc{Max-CSP}}}}
\newcommand{\gnn}{\textsc{GNN}}
\newcommand{\network}{\mbox{RUN-CSP}}
\newcommand{\er}{Erd\H{o}s-R\'{e}nyi}
\newcommand{\np}{\textsf{NP}}

\begin{document}

  \maketitle

  \begin{abstract}
  Many combinatorial optimization problems can be phrased in the language of constraint satisfaction problems. We introduce a graph neural network architecture for solving such optimization problems. The architecture is generic; it works for all binary constraint satisfaction problems. Training is unsupervised, and it is sufficient to train on relatively small instances; the resulting networks perform well on much larger instances (at least 10-times larger).
  We experimentally evaluate our approach for a variety of problems, including Maximum Cut and Maximum Independent Set. Despite being generic, we show that our approach matches or surpasses most greedy and semi-definite programming based algorithms and sometimes even outperforms state-of-the-art heuristics for the specific problems.
  \end{abstract}

\section{Introduction}

Constraint satisfaction is a general framework for casting combinatorial search and optimization problems; many well-known \np{}-complete problems, for example, $k$-colorability, Boolean satisfiability and maximum cut can be modeled as constraint satisfaction problems (\csp{}s).
Our focus is on the optimization version of constraint satisfaction, usually referred to as maximum constraint satisfaction (\textsc{Max-CSP}), where the objective is to satisfy as many constraints of a given instance as possible.
There is a long tradition of designing exact and heuristic algorithms for all kinds of \csp{}s.
Our work should be seen in the context of a recently renewed interest in heuristics for \np{}-hard combinatorial problems based on neural networks, mostly graph neural networks (for example, \cite{selsam2018learning,lemos2019graph,prates2019learning}).

We present a generic graph neural network (GNN) based architecture called \network{} (Recurrent Unsupervised Neural Network for Constraint Satisfaction Problems) with the following key features:
\begin{description}
  \item[Unsupervised:]
    Training is completely unsupervised and just requires a set of instances of the problem.
  \item[Scalable:] 
    Networks trained on small instances achieve good results on much larger inputs.
\item[Generic:]
    The architecture is generic and can learn to find approximate solutions for any binary \mcsp{}.
\end{description}
We focus on binary CSPs, where each constraint involves only two variables. 
This is no severe restriction, because every CSP can be transformed into an equivalent binary CSP (see \cite{dec03}).

To solve \textsc{Max-CSP}s, we train a graph neural network which we view as a message passing protocol. 
The protocol is executed on a graph with nodes for all variables of the instance and edges for all constraints.
Associated with each node are two states, a short-term state and a long-term state, which are both vectors of some fixed length.
The exchanged messages are linear functions of the short-term states.
We update the internal states using an LSTM (Long Short-Term Memory) cell for each variable and share the parameters of the internal functions over all variables. 
Finally, we extract probabilities for the possible values of each variable from its short-term state through a linear function combined with softmax activation. 
The para\-meters of the LSTM, message generation, and readout function are learned.
Since all parameters are shared over all variables, we can apply the model to instances of arbitrary size\footnote{Our Tensorflow implementation of \network{} is available at https://github.com/RUNCSP/RUN-CSP.}.

For training, we design a loss function that rewards solutions with many satisfied constraints.
Effectively, through the choice of the loss function, we train our networks to satisfy the maximum number of constraints.
Our focus on the optimization problem \textsc{Max-CSP} rather than the decision problem allows us to train unsupervised. 
This is a major point of distinction between our work and recent neural approaches to Boolean satisfiability \cite{selsam2018learning} and the coloring problem \cite{lemos2019graph}. 
Both approaches require supervised training and output a prediction for satisfiability or coloring number.
In contrast, our framework returns an (approximately optimal) variable assignment.

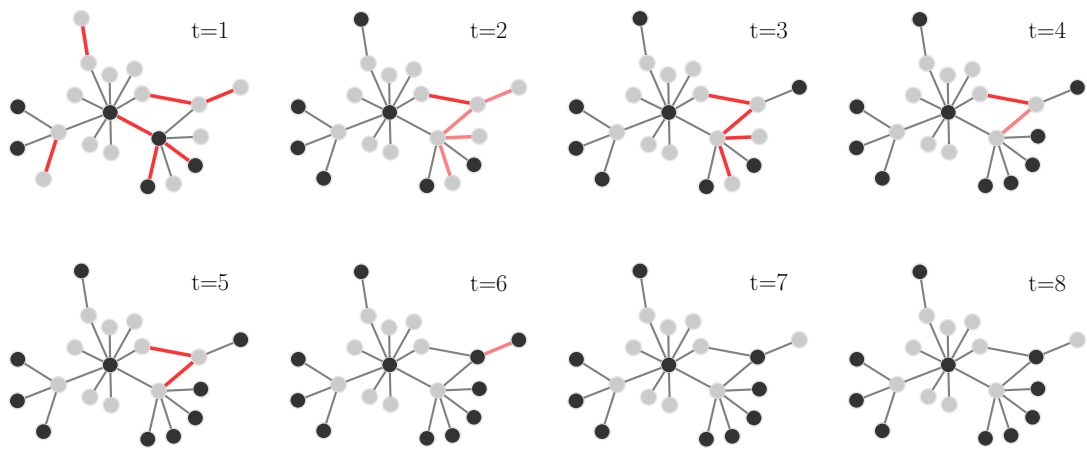
\begin{figure}[t]

\begin{minipage}{0.25\columnwidth}
  \resizebox{0.9\textwidth}{!}{
  \begin{tikzpicture}
\node[font=\LARGE] at (5.0,4.8) {t=1};
\Vertex[x=2.535,y=2.755,size=0.4,color=black,opacity=0.8,style={draw opacity=0.1}]{0}
\Vertex[x=3.734,y=2.109,size=0.4,color=black,opacity=0.8,style={draw opacity=0.1}]{1}
\Vertex[x=3.135,y=3.834,size=0.4,color=lightgray,opacity=0.8,style={draw opacity=0.1}]{2}
\Vertex[x=2.001,y=3.983,size=0.4,color=lightgray,opacity=0.8,style={draw opacity=0.1}]{3}
\Vertex[x=2.520,y=3.689,size=0.4,color=lightgray,opacity=0.8,style={draw opacity=0.1}]{4}
\Vertex[x=1.664,y=3.187,size=0.4,color=lightgray,opacity=0.8,style={draw opacity=0.1}]{5}
\Vertex[x=4.722,y=2.951,size=0.4,color=lightgray,opacity=0.8,style={draw opacity=0.1}]{6}
\Vertex[x=1.261,y=2.263,size=0.4,color=lightgray,opacity=0.8,style={draw opacity=0.1}]{7}
\Vertex[x=5.750,y=3.381,size=0.4,color=lightgray,opacity=0.8,style={draw opacity=0.1}]{8}
\Vertex[x=0.260,y=2.896,size=0.4,color=black,opacity=0.8,style={draw opacity=0.1}]{9}
\Vertex[x=0.250,y=1.864,size=0.4,color=black,opacity=0.8,style={draw opacity=0.1}]{10}
\Vertex[x=1.826,y=5.088,size=0.4,color=lightgray,opacity=0.8,style={draw opacity=0.1}]{11}
\Vertex[x=2.037,y=1.957,size=0.4,color=lightgray,opacity=0.8,style={draw opacity=0.1}]{12}
\Vertex[x=3.461,y=0.912,size=0.4,color=black,opacity=0.8,style={draw opacity=0.1}]{13}
\Vertex[x=0.892,y=1.100,size=0.4,color=lightgray,opacity=0.8,style={draw opacity=0.1}]{14}
\Vertex[x=2.558,y=1.759,size=0.4,color=lightgray,opacity=0.8,style={draw opacity=0.1}]{15}
\Vertex[x=4.097,y=0.984,size=0.4,color=lightgray,opacity=0.8,style={draw opacity=0.1}]{16}
\Vertex[x=4.633,y=1.439,size=0.4,color=black,opacity=0.8,style={draw opacity=0.1}]{17}
\Vertex[x=3.323,y=3.215,size=0.4,color=lightgray,opacity=0.8,style={draw opacity=0.1}]{18}
\Vertex[x=4.767,y=2.155,size=0.4,color=lightgray,opacity=0.8,style={draw opacity=0.1}]{19}
\Edge[,lw=2.5,color=red,opacity=0.8](0)(1)
\Edge[,lw=1.5,color=black,opacity=0.5](0)(2)
\Edge[,lw=1.5,color=black,opacity=0.5](0)(3)
\Edge[,lw=1.5,color=black,opacity=0.5](0)(4)
\Edge[,lw=1.5,color=black,opacity=0.5](0)(5)
\Edge[,lw=1.5,color=black,opacity=0.5](0)(7)
\Edge[,lw=1.5,color=black,opacity=0.5](0)(12)
\Edge[,lw=1.5,color=black,opacity=0.5](0)(15)
\Edge[,lw=1.5,color=black,opacity=0.5](0)(18)
\Edge[,lw=1.5,color=black,opacity=0.5](1)(6)
\Edge[,lw=2.5,color=red,opacity=0.8](1)(13)
\Edge[,lw=1.5,color=black,opacity=0.5](1)(16)
\Edge[,lw=2.5,color=red,opacity=0.8](1)(17)
\Edge[,lw=1.5,color=black,opacity=0.5](1)(19)
\Edge[,lw=2.5,color=red,opacity=0.8](3)(11)
\Edge[,lw=2.5,color=red,opacity=0.8](6)(8)
\Edge[,lw=1.5,color=black,opacity=0.5](7)(9)
\Edge[,lw=1.5,color=black,opacity=0.5](7)(10)
\Edge[,lw=2.5,color=red,opacity=0.8](7)(14)
\Edge[,lw=2.5,color=red,opacity=0.8](6)(18)
\end{tikzpicture}
 }
\end{minipage}\begin{minipage}{0.25\columnwidth}
  \resizebox{0.9\textwidth}{!}{
    \begin{tikzpicture}
\clip (0,0) rectangle (6,6);
\node[font=\LARGE] at (5.0,4.8) {t=2};
\Vertex[x=2.535,y=2.755,size=0.4,color=black,opacity=0.8,style={draw opacity=0.1}]{0}
\Vertex[x=3.734,y=2.109,size=0.4,color=lightgray,opacity=0.8,style={draw opacity=0.1}]{1}
\Vertex[x=3.135,y=3.834,size=0.4,color=lightgray,opacity=0.8,style={draw opacity=0.1}]{2}
\Vertex[x=2.001,y=3.983,size=0.4,color=lightgray,opacity=0.8,style={draw opacity=0.1}]{3}
\Vertex[x=2.520,y=3.689,size=0.4,color=lightgray,opacity=0.8,style={draw opacity=0.1}]{4}
\Vertex[x=1.664,y=3.187,size=0.4,color=lightgray,opacity=0.8,style={draw opacity=0.1}]{5}
\Vertex[x=4.722,y=2.951,size=0.4,color=lightgray,opacity=0.8,style={draw opacity=0.1}]{6}
\Vertex[x=1.261,y=2.263,size=0.4,color=lightgray,opacity=0.8,style={draw opacity=0.1}]{7}
\Vertex[x=5.750,y=3.381,size=0.4,color=lightgray,opacity=0.8,style={draw opacity=0.1}]{8}
\Vertex[x=0.260,y=2.896,size=0.4,color=black,opacity=0.8,style={draw opacity=0.1}]{9}
\Vertex[x=0.250,y=1.864,size=0.4,color=black,opacity=0.8,style={draw opacity=0.1}]{10}
\Vertex[x=1.826,y=5.088,size=0.4,color=black,opacity=0.8,style={draw opacity=0.1}]{11}
\Vertex[x=2.037,y=1.957,size=0.4,color=lightgray,opacity=0.8,style={draw opacity=0.1}]{12}
\Vertex[x=3.461,y=0.912,size=0.4,color=black,opacity=0.8,style={draw opacity=0.1}]{13}
\Vertex[x=0.892,y=1.100,size=0.4,color=black,opacity=0.8,style={draw opacity=0.1}]{14}
\Vertex[x=2.558,y=1.759,size=0.4,color=lightgray,opacity=0.8,style={draw opacity=0.1}]{15}
\Vertex[x=4.097,y=0.984,size=0.4,color=lightgray,opacity=0.8,style={draw opacity=0.1}]{16}
\Vertex[x=4.633,y=1.439,size=0.4,color=black,opacity=0.8,style={draw opacity=0.1}]{17}
\Vertex[x=3.323,y=3.215,size=0.4,color=lightgray,opacity=0.8,style={draw opacity=0.1}]{18}
\Vertex[x=4.767,y=2.155,size=0.4,color=lightgray,opacity=0.8,style={draw opacity=0.1}]{19}
\Edge[,lw=1.5,color=black,opacity=0.5](0)(1)
\Edge[,lw=1.5,color=black,opacity=0.5](0)(2)
\Edge[,lw=1.5,color=black,opacity=0.5](0)(3)
\Edge[,lw=1.5,color=black,opacity=0.5](0)(4)
\Edge[,lw=1.5,color=black,opacity=0.5](0)(5)
\Edge[,lw=1.5,color=black,opacity=0.5](0)(7)
\Edge[,lw=1.5,color=black,opacity=0.5](0)(12)
\Edge[,lw=1.5,color=black,opacity=0.5](0)(15)
\Edge[,lw=1.5,color=black,opacity=0.5](0)(18)
\Edge[,lw=2.5,color=red,opacity=0.5](1)(6)
\Edge[,lw=1.5,color=black,opacity=0.5](1)(13)
\Edge[,lw=2.5,color=red,opacity=0.5](1)(16)
\Edge[,lw=1.5,color=black,opacity=0.5](1)(17)
\Edge[,lw=2.5,color=red,opacity=0.5](1)(19)
\Edge[,lw=1.5,color=black,opacity=0.5](3)(11)
\Edge[,lw=2.5,color=red,opacity=0.5](6)(8)
\Edge[,lw=1.5,color=black,opacity=0.5](7)(9)
\Edge[,lw=1.5,color=black,opacity=0.5](7)(10)
\Edge[,lw=1.5,color=black,opacity=0.5](7)(14)
\Edge[,lw=2.5,color=red,opacity=0.8](6)(18)
\end{tikzpicture}
 }
\end{minipage}\begin{minipage}{0.25\columnwidth}
	\resizebox{0.9\textwidth}{!}{
		\begin{tikzpicture}
\node[font=\LARGE] at (5.0,4.8) {t=3};
\Vertex[x=2.535,y=2.755,size=0.4,color=black,opacity=0.8,style={draw opacity=0.1}]{0}
\Vertex[x=3.734,y=2.109,size=0.4,color=lightgray,opacity=0.8,style={draw opacity=0.1}]{1}
\Vertex[x=3.135,y=3.834,size=0.4,color=lightgray,opacity=0.8,style={draw opacity=0.1}]{2}
\Vertex[x=2.001,y=3.983,size=0.4,color=lightgray,opacity=0.8,style={draw opacity=0.1}]{3}
\Vertex[x=2.520,y=3.689,size=0.4,color=lightgray,opacity=0.8,style={draw opacity=0.1}]{4}
\Vertex[x=1.664,y=3.187,size=0.4,color=lightgray,opacity=0.8,style={draw opacity=0.1}]{5}
\Vertex[x=4.722,y=2.951,size=0.4,color=lightgray,opacity=0.8,style={draw opacity=0.1}]{6}
\Vertex[x=1.261,y=2.263,size=0.4,color=lightgray,opacity=0.8,style={draw opacity=0.1}]{7}
\Vertex[x=5.750,y=3.381,size=0.4,color=black,opacity=0.8,style={draw opacity=0.1}]{8}
\Vertex[x=0.260,y=2.896,size=0.4,color=black,opacity=0.8,style={draw opacity=0.1}]{9}
\Vertex[x=0.250,y=1.864,size=0.4,color=black,opacity=0.8,style={draw opacity=0.1}]{10}
\Vertex[x=1.826,y=5.088,size=0.4,color=black,opacity=0.8,style={draw opacity=0.1}]{11}
\Vertex[x=2.037,y=1.957,size=0.4,color=lightgray,opacity=0.8,style={draw opacity=0.1}]{12}
\Vertex[x=3.461,y=0.912,size=0.4,color=black,opacity=0.8,style={draw opacity=0.1}]{13}
\Vertex[x=0.892,y=1.100,size=0.4,color=black,opacity=0.8,style={draw opacity=0.1}]{14}
\Vertex[x=2.558,y=1.759,size=0.4,color=lightgray,opacity=0.8,style={draw opacity=0.1}]{15}
\Vertex[x=4.097,y=0.984,size=0.4,color=lightgray,opacity=0.8,style={draw opacity=0.1}]{16}
\Vertex[x=4.633,y=1.439,size=0.4,color=black,opacity=0.8,style={draw opacity=0.1}]{17}
\Vertex[x=3.323,y=3.215,size=0.4,color=lightgray,opacity=0.8,style={draw opacity=0.1}]{18}
\Vertex[x=4.767,y=2.155,size=0.4,color=lightgray,opacity=0.8,style={draw opacity=0.1}]{19}
\Edge[,lw=1.5,color=black,opacity=0.5](0)(1)
\Edge[,lw=1.5,color=black,opacity=0.5](0)(2)
\Edge[,lw=1.5,color=black,opacity=0.5](0)(3)
\Edge[,lw=1.5,color=black,opacity=0.5](0)(4)
\Edge[,lw=1.5,color=black,opacity=0.5](0)(5)
\Edge[,lw=1.5,color=black,opacity=0.5](0)(7)
\Edge[,lw=1.5,color=black,opacity=0.5](0)(12)
\Edge[,lw=1.5,color=black,opacity=0.5](0)(15)
\Edge[,lw=1.5,color=black,opacity=0.5](0)(18)
\Edge[,lw=2.5,color=red,opacity=0.8](1)(6)
\Edge[,lw=1.5,color=black,opacity=0.5](1)(13)
\Edge[,lw=2.5,color=red,opacity=0.8](1)(16)
\Edge[,lw=1.5,color=black,opacity=0.5](1)(17)
\Edge[,lw=2.5,color=red,opacity=0.8](1)(19)
\Edge[,lw=1.5,color=black,opacity=0.5](3)(11)
\Edge[,lw=1.5,color=black,opacity=0.5](6)(8)
\Edge[,lw=1.5,color=black,opacity=0.5](7)(9)
\Edge[,lw=1.5,color=black,opacity=0.5](7)(10)
\Edge[,lw=1.5,color=black,opacity=0.5](7)(14)
\Edge[,lw=2.5,color=red,opacity=0.8](6)(18)
\end{tikzpicture}
 }
\end{minipage}\begin{minipage}{0.25\columnwidth}
	\resizebox{0.9\textwidth}{!}{
		\begin{tikzpicture}
\clip (0,0) rectangle (6,6);
\node[font=\LARGE] at (5.0,4.8) {t=4};
\Vertex[x=2.535,y=2.755,size=0.4,color=black,opacity=0.8,style={draw opacity=0.1}]{0}
\Vertex[x=3.734,y=2.109,size=0.4,color=lightgray,opacity=0.8,style={draw opacity=0.1}]{1}
\Vertex[x=3.135,y=3.834,size=0.4,color=lightgray,opacity=0.8,style={draw opacity=0.1}]{2}
\Vertex[x=2.001,y=3.983,size=0.4,color=lightgray,opacity=0.8,style={draw opacity=0.1}]{3}
\Vertex[x=2.520,y=3.689,size=0.4,color=lightgray,opacity=0.8,style={draw opacity=0.1}]{4}
\Vertex[x=1.664,y=3.187,size=0.4,color=lightgray,opacity=0.8,style={draw opacity=0.1}]{5}
\Vertex[x=4.722,y=2.951,size=0.4,color=lightgray,opacity=0.8,style={draw opacity=0.1}]{6}
\Vertex[x=1.261,y=2.263,size=0.4,color=lightgray,opacity=0.8,style={draw opacity=0.1}]{7}
\Vertex[x=5.750,y=3.381,size=0.4,color=black,opacity=0.8,style={draw opacity=0.1}]{8}
\Vertex[x=0.260,y=2.896,size=0.4,color=black,opacity=0.8,style={draw opacity=0.1}]{9}
\Vertex[x=0.250,y=1.864,size=0.4,color=black,opacity=0.8,style={draw opacity=0.1}]{10}
\Vertex[x=1.826,y=5.088,size=0.4,color=black,opacity=0.8,style={draw opacity=0.1}]{11}
\Vertex[x=2.037,y=1.957,size=0.4,color=lightgray,opacity=0.8,style={draw opacity=0.1}]{12}
\Vertex[x=3.461,y=0.912,size=0.4,color=black,opacity=0.8,style={draw opacity=0.1}]{13}
\Vertex[x=0.892,y=1.100,size=0.4,color=black,opacity=0.8,style={draw opacity=0.1}]{14}
\Vertex[x=2.558,y=1.759,size=0.4,color=lightgray,opacity=0.8,style={draw opacity=0.1}]{15}
\Vertex[x=4.097,y=0.984,size=0.4,color=black,opacity=0.8,style={draw opacity=0.1}]{16}
\Vertex[x=4.633,y=1.439,size=0.4,color=black,opacity=0.8,style={draw opacity=0.1}]{17}
\Vertex[x=3.323,y=3.215,size=0.4,color=lightgray,opacity=0.8,style={draw opacity=0.1}]{18}
\Vertex[x=4.767,y=2.155,size=0.4,color=black,opacity=0.8,style={draw opacity=0.1}]{19}
\Edge[,lw=1.5,color=black,opacity=0.5](0)(1)
\Edge[,lw=1.5,color=black,opacity=0.5](0)(2)
\Edge[,lw=1.5,color=black,opacity=0.5](0)(3)
\Edge[,lw=1.5,color=black,opacity=0.5](0)(4)
\Edge[,lw=1.5,color=black,opacity=0.5](0)(5)
\Edge[,lw=1.5,color=black,opacity=0.5](0)(7)
\Edge[,lw=1.5,color=black,opacity=0.5](0)(12)
\Edge[,lw=1.5,color=black,opacity=0.5](0)(15)
\Edge[,lw=1.5,color=black,opacity=0.5](0)(18)
\Edge[,lw=2.5,color=red,opacity=0.5](1)(6)
\Edge[,lw=1.5,color=black,opacity=0.5](1)(13)
\Edge[,lw=1.5,color=black,opacity=0.5](1)(16)
\Edge[,lw=1.5,color=black,opacity=0.5](1)(17)
\Edge[,lw=1.5,color=black,opacity=0.5](1)(19)
\Edge[,lw=1.5,color=black,opacity=0.5](3)(11)
\Edge[,lw=1.5,color=black,opacity=0.5](6)(8)
\Edge[,lw=1.5,color=black,opacity=0.5](7)(9)
\Edge[,lw=1.5,color=black,opacity=0.5](7)(10)
\Edge[,lw=1.5,color=black,opacity=0.5](7)(14)
\Edge[,lw=2.5,color=red,opacity=0.8](6)(18)
\end{tikzpicture}
 }
\end{minipage}\\
\smallskip\\
\begin{minipage}{0.25\columnwidth}
  \resizebox{0.9\textwidth}{!}{

\begin{tikzpicture}
\node[font=\LARGE] at (5.0,4.8) {t=5};
\Vertex[x=2.535,y=2.755,size=0.4,color=black,opacity=0.8,style={draw opacity=0.1}]{0}
\Vertex[x=3.734,y=2.109,size=0.4,color=lightgray,opacity=0.8,style={draw opacity=0.1}]{1}
\Vertex[x=3.135,y=3.834,size=0.4,color=lightgray,opacity=0.8,style={draw opacity=0.1}]{2}
\Vertex[x=2.001,y=3.983,size=0.4,color=lightgray,opacity=0.8,style={draw opacity=0.1}]{3}
\Vertex[x=2.520,y=3.689,size=0.4,color=lightgray,opacity=0.8,style={draw opacity=0.1}]{4}
\Vertex[x=1.664,y=3.187,size=0.4,color=lightgray,opacity=0.8,style={draw opacity=0.1}]{5}
\Vertex[x=4.722,y=2.951,size=0.4,color=lightgray,opacity=0.8,style={draw opacity=0.1}]{6}
\Vertex[x=1.261,y=2.263,size=0.4,color=lightgray,opacity=0.8,style={draw opacity=0.1}]{7}
\Vertex[x=5.750,y=3.381,size=0.4,color=black,opacity=0.8,style={draw opacity=0.1}]{8}
\Vertex[x=0.260,y=2.896,size=0.4,color=black,opacity=0.8,style={draw opacity=0.1}]{9}
\Vertex[x=0.250,y=1.864,size=0.4,color=black,opacity=0.8,style={draw opacity=0.1}]{10}
\Vertex[x=1.826,y=5.088,size=0.4,color=black,opacity=0.8,style={draw opacity=0.1}]{11}
\Vertex[x=2.037,y=1.957,size=0.4,color=lightgray,opacity=0.8,style={draw opacity=0.1}]{12}
\Vertex[x=3.461,y=0.912,size=0.4,color=black,opacity=0.8,style={draw opacity=0.1}]{13}
\Vertex[x=0.892,y=1.100,size=0.4,color=black,opacity=0.8,style={draw opacity=0.1}]{14}
\Vertex[x=2.558,y=1.759,size=0.4,color=lightgray,opacity=0.8,style={draw opacity=0.1}]{15}
\Vertex[x=4.097,y=0.984,size=0.4,color=black,opacity=0.8,style={draw opacity=0.1}]{16}
\Vertex[x=4.633,y=1.439,size=0.4,color=black,opacity=0.8,style={draw opacity=0.1}]{17}
\Vertex[x=3.323,y=3.215,size=0.4,color=lightgray,opacity=0.8,style={draw opacity=0.1}]{18}
\Vertex[x=4.767,y=2.155,size=0.4,color=black,opacity=0.8,style={draw opacity=0.1}]{19}
\Edge[,lw=1.5,color=black,opacity=0.5](0)(1)
\Edge[,lw=1.5,color=black,opacity=0.5](0)(2)
\Edge[,lw=1.5,color=black,opacity=0.5](0)(3)
\Edge[,lw=1.5,color=black,opacity=0.5](0)(4)
\Edge[,lw=1.5,color=black,opacity=0.5](0)(5)
\Edge[,lw=1.5,color=black,opacity=0.5](0)(7)
\Edge[,lw=1.5,color=black,opacity=0.5](0)(12)
\Edge[,lw=1.5,color=black,opacity=0.5](0)(15)
\Edge[,lw=1.5,color=black,opacity=0.5](0)(18)
\Edge[,lw=2.5,color=red,opacity=0.8](1)(6)
\Edge[,lw=1.5,color=black,opacity=0.5](1)(13)
\Edge[,lw=1.5,color=black,opacity=0.5](1)(16)
\Edge[,lw=1.5,color=black,opacity=0.5](1)(17)
\Edge[,lw=1.5,color=black,opacity=0.5](1)(19)
\Edge[,lw=1.5,color=black,opacity=0.5](3)(11)
\Edge[,lw=1.5,color=black,opacity=0.5](6)(8)
\Edge[,lw=1.5,color=black,opacity=0.5](7)(9)
\Edge[,lw=1.5,color=black,opacity=0.5](7)(10)
\Edge[,lw=1.5,color=black,opacity=0.5](7)(14)
\Edge[,lw=2.5,color=red,opacity=0.8](6)(18)
\end{tikzpicture}
 }
\end{minipage}\begin{minipage}{0.25\columnwidth}
  \resizebox{0.9\textwidth}{!}{
    \begin{tikzpicture}
\clip (0,0) rectangle (6,6);
\node[font=\LARGE] at (5.0,4.8) {t=6};
\Vertex[x=2.535,y=2.755,size=0.4,color=black,opacity=0.8,style={draw opacity=0.1}]{0}
\Vertex[x=3.734,y=2.109,size=0.4,color=lightgray,opacity=0.8,style={draw opacity=0.1}]{1}
\Vertex[x=3.135,y=3.834,size=0.4,color=lightgray,opacity=0.8,style={draw opacity=0.1}]{2}
\Vertex[x=2.001,y=3.983,size=0.4,color=lightgray,opacity=0.8,style={draw opacity=0.1}]{3}
\Vertex[x=2.520,y=3.689,size=0.4,color=lightgray,opacity=0.8,style={draw opacity=0.1}]{4}
\Vertex[x=1.664,y=3.187,size=0.4,color=lightgray,opacity=0.8,style={draw opacity=0.1}]{5}
\Vertex[x=4.722,y=2.951,size=0.4,color=black,opacity=0.8,style={draw opacity=0.1}]{6}
\Vertex[x=1.261,y=2.263,size=0.4,color=lightgray,opacity=0.8,style={draw opacity=0.1}]{7}
\Vertex[x=5.750,y=3.381,size=0.4,color=black,opacity=0.8,style={draw opacity=0.1}]{8}
\Vertex[x=0.260,y=2.896,size=0.4,color=black,opacity=0.8,style={draw opacity=0.1}]{9}
\Vertex[x=0.250,y=1.864,size=0.4,color=black,opacity=0.8,style={draw opacity=0.1}]{10}
\Vertex[x=1.826,y=5.088,size=0.4,color=black,opacity=0.8,style={draw opacity=0.1}]{11}
\Vertex[x=2.037,y=1.957,size=0.4,color=lightgray,opacity=0.8,style={draw opacity=0.1}]{12}
\Vertex[x=3.461,y=0.912,size=0.4,color=black,opacity=0.8,style={draw opacity=0.1}]{13}
\Vertex[x=0.892,y=1.100,size=0.4,color=black,opacity=0.8,style={draw opacity=0.1}]{14}
\Vertex[x=2.558,y=1.759,size=0.4,color=lightgray,opacity=0.8,style={draw opacity=0.1}]{15}
\Vertex[x=4.097,y=0.984,size=0.4,color=black,opacity=0.8,style={draw opacity=0.1}]{16}
\Vertex[x=4.633,y=1.439,size=0.4,color=black,opacity=0.8,style={draw opacity=0.1}]{17}
\Vertex[x=3.323,y=3.215,size=0.4,color=lightgray,opacity=0.8,style={draw opacity=0.1}]{18}
\Vertex[x=4.767,y=2.155,size=0.4,color=black,opacity=0.8,style={draw opacity=0.1}]{19}
\Edge[,lw=1.5,color=black,opacity=0.5](0)(1)
\Edge[,lw=1.5,color=black,opacity=0.5](0)(2)
\Edge[,lw=1.5,color=black,opacity=0.5](0)(3)
\Edge[,lw=1.5,color=black,opacity=0.5](0)(4)
\Edge[,lw=1.5,color=black,opacity=0.5](0)(5)
\Edge[,lw=1.5,color=black,opacity=0.5](0)(7)
\Edge[,lw=1.5,color=black,opacity=0.5](0)(12)
\Edge[,lw=1.5,color=black,opacity=0.5](0)(15)
\Edge[,lw=1.5,color=black,opacity=0.5](0)(18)
\Edge[,lw=1.5,color=black,opacity=0.5](1)(6)
\Edge[,lw=1.5,color=black,opacity=0.5](1)(13)
\Edge[,lw=1.5,color=black,opacity=0.5](1)(16)
\Edge[,lw=1.5,color=black,opacity=0.5](1)(17)
\Edge[,lw=1.5,color=black,opacity=0.5](1)(19)
\Edge[,lw=1.5,color=black,opacity=0.5](3)(11)
\Edge[,lw=2.5,color=red,opacity=0.5](6)(8)
\Edge[,lw=1.5,color=black,opacity=0.5](7)(9)
\Edge[,lw=1.5,color=black,opacity=0.5](7)(10)
\Edge[,lw=1.5,color=black,opacity=0.5](7)(14)
\Edge[,lw=1.5,color=black,opacity=0.5](6)(18)
\end{tikzpicture}
 }
\end{minipage}\begin{minipage}{0.25\columnwidth}
	\resizebox{0.9\textwidth}{!}{
		\begin{tikzpicture}
\node[font=\LARGE] at (5.0,4.8) {t=7};
\Vertex[x=2.535,y=2.755,size=0.4,color=black,opacity=0.8,style={draw opacity=0.1}]{0}
\Vertex[x=3.734,y=2.109,size=0.4,color=lightgray,opacity=0.8,style={draw opacity=0.1}]{1}
\Vertex[x=3.135,y=3.834,size=0.4,color=lightgray,opacity=0.8,style={draw opacity=0.1}]{2}
\Vertex[x=2.001,y=3.983,size=0.4,color=lightgray,opacity=0.8,style={draw opacity=0.1}]{3}
\Vertex[x=2.520,y=3.689,size=0.4,color=lightgray,opacity=0.8,style={draw opacity=0.1}]{4}
\Vertex[x=1.664,y=3.187,size=0.4,color=lightgray,opacity=0.8,style={draw opacity=0.1}]{5}
\Vertex[x=4.722,y=2.951,size=0.4,color=black,opacity=0.8,style={draw opacity=0.1}]{6}
\Vertex[x=1.261,y=2.263,size=0.4,color=lightgray,opacity=0.8,style={draw opacity=0.1}]{7}
\Vertex[x=5.750,y=3.381,size=0.4,color=lightgray,opacity=0.8,style={draw opacity=0.1}]{8}
\Vertex[x=0.260,y=2.896,size=0.4,color=black,opacity=0.8,style={draw opacity=0.1}]{9}
\Vertex[x=0.250,y=1.864,size=0.4,color=black,opacity=0.8,style={draw opacity=0.1}]{10}
\Vertex[x=1.826,y=5.088,size=0.4,color=black,opacity=0.8,style={draw opacity=0.1}]{11}
\Vertex[x=2.037,y=1.957,size=0.4,color=lightgray,opacity=0.8,style={draw opacity=0.1}]{12}
\Vertex[x=3.461,y=0.912,size=0.4,color=black,opacity=0.8,style={draw opacity=0.1}]{13}
\Vertex[x=0.892,y=1.100,size=0.4,color=black,opacity=0.8,style={draw opacity=0.1}]{14}
\Vertex[x=2.558,y=1.759,size=0.4,color=lightgray,opacity=0.8,style={draw opacity=0.1}]{15}
\Vertex[x=4.097,y=0.984,size=0.4,color=black,opacity=0.8,style={draw opacity=0.1}]{16}
\Vertex[x=4.633,y=1.439,size=0.4,color=black,opacity=0.8,style={draw opacity=0.1}]{17}
\Vertex[x=3.323,y=3.215,size=0.4,color=lightgray,opacity=0.8,style={draw opacity=0.1}]{18}
\Vertex[x=4.767,y=2.155,size=0.4,color=black,opacity=0.8,style={draw opacity=0.1}]{19}
\Edge[,lw=1.5,color=black,opacity=0.5](0)(1)
\Edge[,lw=1.5,color=black,opacity=0.5](0)(2)
\Edge[,lw=1.5,color=black,opacity=0.5](0)(3)
\Edge[,lw=1.5,color=black,opacity=0.5](0)(4)
\Edge[,lw=1.5,color=black,opacity=0.5](0)(5)
\Edge[,lw=1.5,color=black,opacity=0.5](0)(7)
\Edge[,lw=1.5,color=black,opacity=0.5](0)(12)
\Edge[,lw=1.5,color=black,opacity=0.5](0)(15)
\Edge[,lw=1.5,color=black,opacity=0.5](0)(18)
\Edge[,lw=1.5,color=black,opacity=0.5](1)(6)
\Edge[,lw=1.5,color=black,opacity=0.5](1)(13)
\Edge[,lw=1.5,color=black,opacity=0.5](1)(16)
\Edge[,lw=1.5,color=black,opacity=0.5](1)(17)
\Edge[,lw=1.5,color=black,opacity=0.5](1)(19)
\Edge[,lw=1.5,color=black,opacity=0.5](3)(11)
\Edge[,lw=1.5,color=black,opacity=0.5](6)(8)
\Edge[,lw=1.5,color=black,opacity=0.5](7)(9)
\Edge[,lw=1.5,color=black,opacity=0.5](7)(10)
\Edge[,lw=1.5,color=black,opacity=0.5](7)(14)
\Edge[,lw=1.5,color=black,opacity=0.5](6)(18)
\end{tikzpicture}
 }
\end{minipage}\begin{minipage}{0.25\columnwidth}
	\resizebox{0.9\textwidth}{!}{
		\begin{tikzpicture}
\clip (0,0) rectangle (6,6);
\node[font=\LARGE] at (5.0,4.8) {t=8};
\Vertex[x=2.535,y=2.755,size=0.4,color=black,opacity=0.8,style={draw opacity=0.1}]{0}
\Vertex[x=3.734,y=2.109,size=0.4,color=lightgray,opacity=0.8,style={draw opacity=0.1}]{1}
\Vertex[x=3.135,y=3.834,size=0.4,color=lightgray,opacity=0.8,style={draw opacity=0.1}]{2}
\Vertex[x=2.001,y=3.983,size=0.4,color=lightgray,opacity=0.8,style={draw opacity=0.1}]{3}
\Vertex[x=2.520,y=3.689,size=0.4,color=lightgray,opacity=0.8,style={draw opacity=0.1}]{4}
\Vertex[x=1.664,y=3.187,size=0.4,color=lightgray,opacity=0.8,style={draw opacity=0.1}]{5}
\Vertex[x=4.722,y=2.951,size=0.4,color=black,opacity=0.8,style={draw opacity=0.1}]{6}
\Vertex[x=1.261,y=2.263,size=0.4,color=lightgray,opacity=0.8,style={draw opacity=0.1}]{7}
\Vertex[x=5.750,y=3.381,size=0.4,color=lightgray,opacity=0.8,style={draw opacity=0.1}]{8}
\Vertex[x=0.260,y=2.896,size=0.4,color=black,opacity=0.8,style={draw opacity=0.1}]{9}
\Vertex[x=0.250,y=1.864,size=0.4,color=black,opacity=0.8,style={draw opacity=0.1}]{10}
\Vertex[x=1.826,y=5.088,size=0.4,color=black,opacity=0.8,style={draw opacity=0.1}]{11}
\Vertex[x=2.037,y=1.957,size=0.4,color=lightgray,opacity=0.8,style={draw opacity=0.1}]{12}
\Vertex[x=3.461,y=0.912,size=0.4,color=black,opacity=0.8,style={draw opacity=0.1}]{13}
\Vertex[x=0.892,y=1.100,size=0.4,color=black,opacity=0.8,style={draw opacity=0.1}]{14}
\Vertex[x=2.558,y=1.759,size=0.4,color=lightgray,opacity=0.8,style={draw opacity=0.1}]{15}
\Vertex[x=4.097,y=0.984,size=0.4,color=black,opacity=0.8,style={draw opacity=0.1}]{16}
\Vertex[x=4.633,y=1.439,size=0.4,color=black,opacity=0.8,style={draw opacity=0.1}]{17}
\Vertex[x=3.323,y=3.215,size=0.4,color=lightgray,opacity=0.8,style={draw opacity=0.1}]{18}
\Vertex[x=4.767,y=2.155,size=0.4,color=black,opacity=0.8,style={draw opacity=0.1}]{19}
\Edge[,lw=1.5,color=black,opacity=0.5](0)(1)
\Edge[,lw=1.5,color=black,opacity=0.5](0)(2)
\Edge[,lw=1.5,color=black,opacity=0.5](0)(3)
\Edge[,lw=1.5,color=black,opacity=0.5](0)(4)
\Edge[,lw=1.5,color=black,opacity=0.5](0)(5)
\Edge[,lw=1.5,color=black,opacity=0.5](0)(7)
\Edge[,lw=1.5,color=black,opacity=0.5](0)(12)
\Edge[,lw=1.5,color=black,opacity=0.5](0)(15)
\Edge[,lw=1.5,color=black,opacity=0.5](0)(18)
\Edge[,lw=1.5,color=black,opacity=0.5](1)(6)
\Edge[,lw=1.5,color=black,opacity=0.5](1)(13)
\Edge[,lw=1.5,color=black,opacity=0.5](1)(16)
\Edge[,lw=1.5,color=black,opacity=0.5](1)(17)
\Edge[,lw=1.5,color=black,opacity=0.5](1)(19)
\Edge[,lw=1.5,color=black,opacity=0.5](3)(11)
\Edge[,lw=1.5,color=black,opacity=0.5](6)(8)
\Edge[,lw=1.5,color=black,opacity=0.5](7)(9)
\Edge[,lw=1.5,color=black,opacity=0.5](7)(10)
\Edge[,lw=1.5,color=black,opacity=0.5](7)(14)
\Edge[,lw=1.5,color=black,opacity=0.5](6)(18)
\end{tikzpicture}
 }
\end{minipage}%
\caption{A maximum cut for a graph found by \network{} in seven iterations. Edges not part of the cut are shown in red.
	}
	\label{fig:maxCutAnimation}
\end{figure}

We experimentally evaluate our approach on the following \np{}-hard problems:
the maximum 2-satisfiability problem (\msat{}), which asks for an assignment maximizing the number of satisfied clauses for a given Boolean formula in 2-conjunctive normal form; 
the maximum cut problem (\mcut{}), which asks for a partition of a graph in two parts such that the number of edges between the parts is maximal (see Figure~\ref{fig:maxCutAnimation});
the 3-colorability problem (\dcol{}), which asks for a 3-coloring of the vertices of a given graph such that the two endvertices of each edge have distinct colors. We also consider the maximum independent set problem \mis{}, which asks for an independent set of maximum cardinality in a given graph.
Strictly speaking, \mis{} is not a maximum constraint satisfaction problem, because its objective is not to maximize the number of satisfied constraints, but to satisfy all constraints while maximizing the number of variables with a certain value. 
We include this problem to demonstrate that our approach can easily be adapted to such related problems.

We demonstrate that our simple generic approach works well for all four problems and matches competitive baselines.
Since our approach is generic for all \textsc{Max-CSP}s, those baselines include other general approaches such as greedy algorithms and semi-definite programming (SDP). 
The latter is particularly relevant, because it is known (under certain complexity theoretic assumptions) that SDP achieves optimal approximation ratios for all \textsc{Max-CSP}s \cite{rag08}.
For \msat{}, our approach even manages to surpass a state-of-the-art heuristic.

Almost all models are trained on quite small training sets consisting of small random instances.
We evaluate those models on unstructured random instances as well as highly structured benchmark instances.
Instance sizes vary from small instances with 100 variables and 200 constraints to medium sized instances with more than 1,000 variables and over 10,000 constraints.
We observe that \network{} is able to generalize well from small instances to instances both smaller and much larger. 
The largest (benchmark) instance we evaluate on has approximately 120,000 constraints, but that instance required the use of large training graphs.
Computations with \network{} are very fast in comparison to many heuristics and profits from modern hardware like GPUs. 
For medium-sized instances with 10,000 constraints the computation takes less than five seconds.

We do not claim that our method is in general competitive with state-of-the-art heuristics for specific problems. 
Furthermore, it has to be distinguished from solvers which in addition to a solution return a guarantee that no better solution exists.
However, we demonstrate that our approach clearly improves on the state-of-the-art for neural methods on small and medium-sized binary CSP instances, while still being completely generic.

\subsection{Related Work}

An early group of papers dates back to the 1980's and uses Hopfield Networks \cite{hopfield1985neural} to approximate \tsp{} and other discrete problems using neural networks.
\citeauthor{hopfield1985neural} use a single-layer neural network with sigmoid activation and apply gradient descent to come up with an approximative solution.
The loss function adopts soft assignments and uses the length of the \tsp{} tour and a term penalizing incorrect tours as loss, hence being unsupervised.
This approach has been extended to $k$-colorability \cite{Dahl1987NeuralNA,takefuji1991artificial,harmanani2010neural,gassen1993graph} and other \csp{}s \cite{adorf1990discrete}.
The loss functions used in some of these approaches are similar to ours.

Newer approaches involve modern machine learning techniques and are usually based on graph neural networks (\gnn{}s).
NeuroSAT \cite{selsam2018learning}, a learned message passing network for predicting satisfiability, reignited the interest in solving \np{}-complete problems with neural networks.
\citeinline{prates2019learning} use \gnn{}s to learn \tsp{} trained on instances of the form $(G,\ell \!\pm\! \varepsilon)$ where $\ell$ is the length of an optimal tour on $G$.
They achieved good results on graphs with up to $40$ nodes.
Using the same idea, \citeinline{lemos2019graph} learned to predict $k$-colorability of graphs scaling to larger graphs and chromatic numbers than seen during training.
\citeinline{yao2019experimental} evaluated the performance of unsupervised \gnn{}s for the \mcut{} problem.
They adapted a \gnn{} architecture by \citeinline{chen2017supervised} to \mcut{} and trained two versions of their network, one through policy gradient descent and the other via a differentiable relaxation of the loss function which both achieved similar results.
\citeinline{amizadeh2018learning} proposed an unsupervised architecture for \textsc{Circuit-SAT}, which predicts satisfying variable assignments for a given formula.
For the \textsf{\#P}-hard weighted model counting problem for DNF formulas, \citeinline{abboud2019learning} applied a \gnn{}-based message passing approach.
For large instances with more than 100,000 nodes, \citeinline{li2018combinatorial} use a \gnn{} to guide a tree search for \mis{} and \citeinline{khalil2017learning} choose the best heuristic for \tsp{} through reinforcement learning.

\section{Method}\label{sec:method}
In this section, we describe our \network{} architecture for \textsc{Max-CSP}s.
Formally, a CSP-instance is a triple $I=(X,D,C)$, where $X$ is a set of variables, $D$ is a domain, and $C$ is a set of constraints of the form $(x_1,\ldots,x_\ell,R)$ for some $R\subseteq D^\ell$.
We only consider binary constraints (with $\ell=2$) in this paper.
A \emph{constraint language} is a finite set $\Gamma$ of relations over some fixed domain $D$, and $I$ is a $\Gamma$-instance if $R\in\Gamma$ for all constraints $(x_1,x_2,R)\in C$.
An \emph{assignment} $\alpha:X\to D$ satisfies a constraint $(x_1,x_2,R)$ if $(\alpha(x_1),\alpha(x_2))\in R$, and it satisfies the instance $I$ if it satisfies all constraints in $C$.
$\textsc{CSP}(\Gamma)$ is the problem of deciding whether a given $\Gamma$-instance has a satisfying assignment{ and finding such an assignment if there is one}.
$\textsc{Max-CSP}(\Gamma)$ is the problem of finding an assignment that satisfies the maximum number of constraints.

For example, an instance of \dcol{} has a variable $x_v$ for each vertex $v$ of the input graph, domain $D=\{1,2,3\}$, and a constraint $(v,w,R^3_{\neq})$ for each edge $vw$ of the graph.
Here $R^3_{\neq}=\{(1,2),(2,1),(1,3),(3,1),(2,3),(3,2)\}$ is the inequality relation on $\{1,2,3\}$.
Thus \dcol{} is $\textsc{CSP}(\{R^3_{\neq}\})$.

\subsection{Architecture}\label{subsec:architecture}

We use a randomized recurrent graph neural network architecture to evaluate a given problem instance using message passing.
For any binary constraint language $\Gamma$ a \network{} network can be trained to approximate  $\textsc{Max-CSP}(\Gamma)$.
Intuitively, our network can be viewed as a trainable communication protocol through which the variables of a given instance can negotiate a value assignment. 
With every variable $x\in X$ we associate a short-term state $s_x^{(t)} \in \RR^k$  and a hidden (long-term) state $h_x^{(t)} \in \RR^k$ which change throughout the message passing iterations $t \in \{0,\ldots,\tmax{}\}$.
The short-term state vector $s_x^{(0)}$ for every variable $x$ is initialized by sampling each value independently from a normal distribution with zero mean and unit variance.
All hidden states $h_x^{(0)}$ are initialized as zero vectors.

Every message passing step uses the same weights and thus we are free to choose the number $\tmax{}\in \NN$ of iterations for which \network{} runs on a given problem instance.
This number may or may not be identical to the number of iterations used for training.
The state size $k$ and the number of iterations used for training $\train{}$ and evaluation $\eval{}$ are the main hyperparameters of our network.

Variables $x$ and $y$ that co-occur in a constraint $c = (x,y,R)$ can exchange messages.
Each message depends on the states $s_x^{(t)},s_y^{(t)}$, the relation $R$ and the order of $x$ and $y$ in the constraint, but not the internal long-term states $h_x^{(t)},h_y^{(t)}$.
The dependence on $R$ allows the network to send different messages whenever the states of $x$ and $y$ correspond to a satisfying or unsatisfying assignment for each constraint.
This dependence implies that we have independent message generation functions for every relation $R$ in the constraint language $\Gamma$.
We therefore require $\Gamma$ to be fixed.
The process of message passing and updating the internal states is repeated $\tmax{}$ times.
We use linear functions to compute the messages. (Experiments showed that more complicated functions did not improve performance while being less stable and less efficient during training.)
Thus, the messaging function for every relation $R$ is defined by a trainable weight matrix $M_{\!R} \in \RR^{2k \times 2k}$ as
\begin{equation}
  S_{\!R}\left(s_x^{(t)}, s_y^{(t)}\right) = M_{\!R} \begin{pmatrix}s_x^{(t)}\\s_y^{(t)}\end{pmatrix}.\\
\end{equation}
The output of $S_R$ are two stacked $k$-dimensional vectors, which represent the messages to $x$ and $y$, respectively.
Since $M_{\!R}$ is in general not symmetric, the generated messages depend on the order of the variables in the constraint.
This behavior is desirable for asymmetric relations. 
For symmetric relations we modify $S_R$ to produce messages independently from the order of variables in $c$. 
In this case we use a smaller weight matrix $M_{\!R} \in \RR^{2k \times k}$ to generate both messages:
\begin{equation*}
S_{\!R}(s_x^{(t)}, s_y^{(t)}) = \left(
  		M_{\!R} \begin{pmatrix}s_x^{(t)}\\s_y^{(t)}\end{pmatrix}, 
  		M_{\!R} \begin{pmatrix}s_y^{(t)}\\s_x^{(t)}\end{pmatrix}
   \right)
\end{equation*}
The two messages can still be different, but the content of each message depends only on the states of the endpoints.

The internal states $h_x$ and $s_x$ are updated by an LSTM cell based on the mean of the received messages.
For a variable $x$ which received the messages $m_1,\dots,m_\ell$ the new states are thus computed by
\begin{equation}
  h_x^{(t+1)}, s_x^{(t+1)} = \text{LSTM}\bigg(h_x^{(t)},\,s_x^{(t)},\,\frac{1}{\ell}\sum_{i=1}^\ell m_i\bigg).
\end{equation}
For every variable $x$ and iteration $t \in \{1,\dots,\tmax{}\}$ the network produces a soft assignment $\phi^{(t)}(x)$ from the state $s_x^{(t)}$.
In our architecture we use $\phi^{(t)}(x) = \text{softmax}(W s^{(t)}_x)$ with $W \in \RR^{d \times k}$ trainable and $d=|D|$. 
In $\phi$, the linear function reduces the dimensionality while the softmax function enforces stochasticity.
The soft assignments $\phi^{(t)}(x)$ can be interpreted as probabilities of a variable $x$ receiving a certain value $v\in D$. 
If the domain $D$ contains only two values, we compute a `probability' $p^{(t)}(x) = \sigma(W s^{(t)}_x)$ for each node with $W \in \mathbb{R}^{1 \times k}$.
The soft assignment is then given by $\varphi^{(t)}(x) = \left(p^{(t)}(x), 1-p^{(t)}(x)\right)$.
To obtain a hard variable assignment $\alpha^{(t)}:X \rightarrow D$, we assign the value with the highest estimated probability in $\phi^{(t)}(x)$ for each variable $x\in X$.
We select the hard assignment with the most satisfied constraints as the final prediction of the network.
This is not necessarily $\alpha^{(\eval{})}$.

Algorithm \ref{Alg:Network} describes the architecture in pseudocode.
Note that the network's output depends on the random initialization of the short-term states $s_x^{(0)}$.
Those states are the basis for all messages sent during inference and thus for the solution found by \network{}.
By applying the network multiple times to the same input and choosing the best solution, we can therefore boost the performance.

\begin{algorithm}[t]
	\begin{algorithmic}
		\STATE {\bfseries Input:} Instance $(X, C)$, $t_\text{max} \in \mathbb{N}$
		\STATE {\bfseries Output:} $(\varphi^{(1)},\dots,\varphi^{(t_\text{max})}), \varphi^{(t)}: X \rightarrow [0, 1]^d$
    \FOR{$x \in X$}
\STATE\COMMENT{random initialization}
			\STATE $s_x^{(0)} \sim \mathcal{N}(0, 1)^k$ 
			\STATE $h_x^{(0)} \coloneqq \bm{0} \in \RR{}^k $\ENDFOR
\FOR{$t \in \{1,...,t_\text{max}\}$}
			\FOR{$c \coloneqq (x, y, R) \in C$}
				\STATE \COMMENT{generate messages}
				\STATE $(m_{c,x}^{(t)},\, m_{c,y}^{(t)}) \coloneqq S_{\!R}(s_x^{(t-1)},\, s_y^{(t-1)})$
			\ENDFOR
\FOR{$x \in X$}
        \STATE \COMMENT{combine messages and update}
				\STATE $r_x^{(t)} \coloneqq \frac{1}{\text{deg}(x)} \sum_{c \in C, x \in c} m_{c,x}^{(t)}$
\STATE $(h_x^{(t)}, s_x^{(t)}) \coloneqq \text{LSTM}(h_x^{(t-1)},\, s_x^{(t-1)},\, r_x^{(t)})$
\STATE $\varphi^{(t)}(x) \coloneqq \softmax(W \cdot s_x^{(t)})$
			\ENDFOR
		\ENDFOR
\end{algorithmic}
  \caption{Network Architecture}
  \label{Alg:Network}
\end{algorithm}

\subsection{Loss Function}

  In the following we derive our loss function used for unsupervised training. 
  Let $I=(X,D,C)$ be a \csp{}-instance.
  Assume without loss of generality that $D=\{1,\ldots,d\}$ for a positive integer $d$. 
  Given $I$, in every iteration our network will produce a soft variable assignment $\phi: X \rightarrow [0,1]^d$, where $\phi(x)$ is stochastic for every $x \in X$. 
  Instead of choosing the value with the maximum probability in $\phi(x)$, we could obtain a hard assignment $\alpha: X \rightarrow D$ by independently sampling a value for each $x\in X$ from the distribution specified by $\phi(x)$. 
  In this case, the probability that any given constraint $(x,y,R) \in C$ is satisfied by $\alpha$ can be expressed by
\begin{equation}
    \Pr_{\alpha \sim \varphi}\left[(\alpha(x),\alpha(y))\in R\right] = \phi(x)^T A_R \, \phi(y)
  \end{equation}
where $A_R \in \{0,1\}^{d \times d}$ is the characteristic matrix of the relation $R$ with $(A_R)_{i,j}=1 \iff (i,j) \in R$. 
  We then aim to minimize the combined negative log-likelihood over all constraints:
\begin{equation} \label{eq:lossFunction}
    \mathcal{L}_\text{CSP}\left(\varphi, I\right) \coloneqq  \frac{1}{|C|}\cdot\!\!\! \sum_{(x,y,R) \in C}\!\!\!\! -\log\left(\phi(x)^T A_R \, \varphi(y)\right)
  \end{equation}
  We combine the loss function $\mathcal L_\text{CSP}$ throughout all iterations with a discount factor $\lambda \in [0,1]$ to get our training objective:
  \begin{equation}
    \mathcal{L}(\{\phi_t\}_{t \leq \train{}},I) \coloneqq 
      \sum_{t = 1}^{\train{}} \lambda^{\train{}-t} \cdot \mathcal{L}_\text{CSP}\left(\varphi^{(t)}, I\right)
  \end{equation}
  This loss function allows us to train unsupervised since it does not depend on any ground truth assignments.
In general, computing optimal solutions for supervised training can easily turn out to be prohibitive; our approach completely avoids such computations.

\medskip
\textit{Remarks:}

(1) In this paper, we focus on binary CSPs. 
For the extension to $\ell$-ary CSPs for some $\ell\ge 3$ (for example $\textsc{3-SAT}$), we note that the generalization of the loss function is straightforward.
Exploring network architectures that can process constraints of higher arity remains future work.

(2) It is also possible to extend the framework to the weighted \mcsp{}s where a weight is associated with each constraint. 
To achieve this, we can replace the averages in the loss function and message collection steps by weighted averages.
Early experiments in that direction show promising results.

\section{Experiments}\label{sec:experiments}

To validate our method empirically, we performed experiments for \msat{}, \mcut{}, \dcol{} and \mis{}.
For all experiments, we used internal states of size $k=128$; state sizes up to $k=1024$ did not increase performance for the tested instances.
We empirically chose to use $\train{}=30$ iterations during training and, unless stated otherwise, $\eval{}=100$ for evaluation.
Especially for larger instances it proved beneficial to use a relatively high $\eval{}$.
In contrast, choosing $\train{}$ too large during training ($\train{}>50$) resulted in unstable training.
During evaluation, we use $64$ parallel runs for each instance and use the best result. 
Further increasing this number mainly increases the runtime but has no real effect on the quality of solutions.
We trained most models with 4,000 instances split into in $400$ batches.
Training is performed for $25$ epochs using the Adam optimizer with default parameters and gradient clipping at a norm of $1.0$. 
The decay over time in our loss function was set to $\lambda=0.95$.
We provide a more detailed overview of our implementation and training configuration in the appendix.

\begin{figure*}[t]
  \begin{center}
    \includegraphics[width=1.0\textwidth]{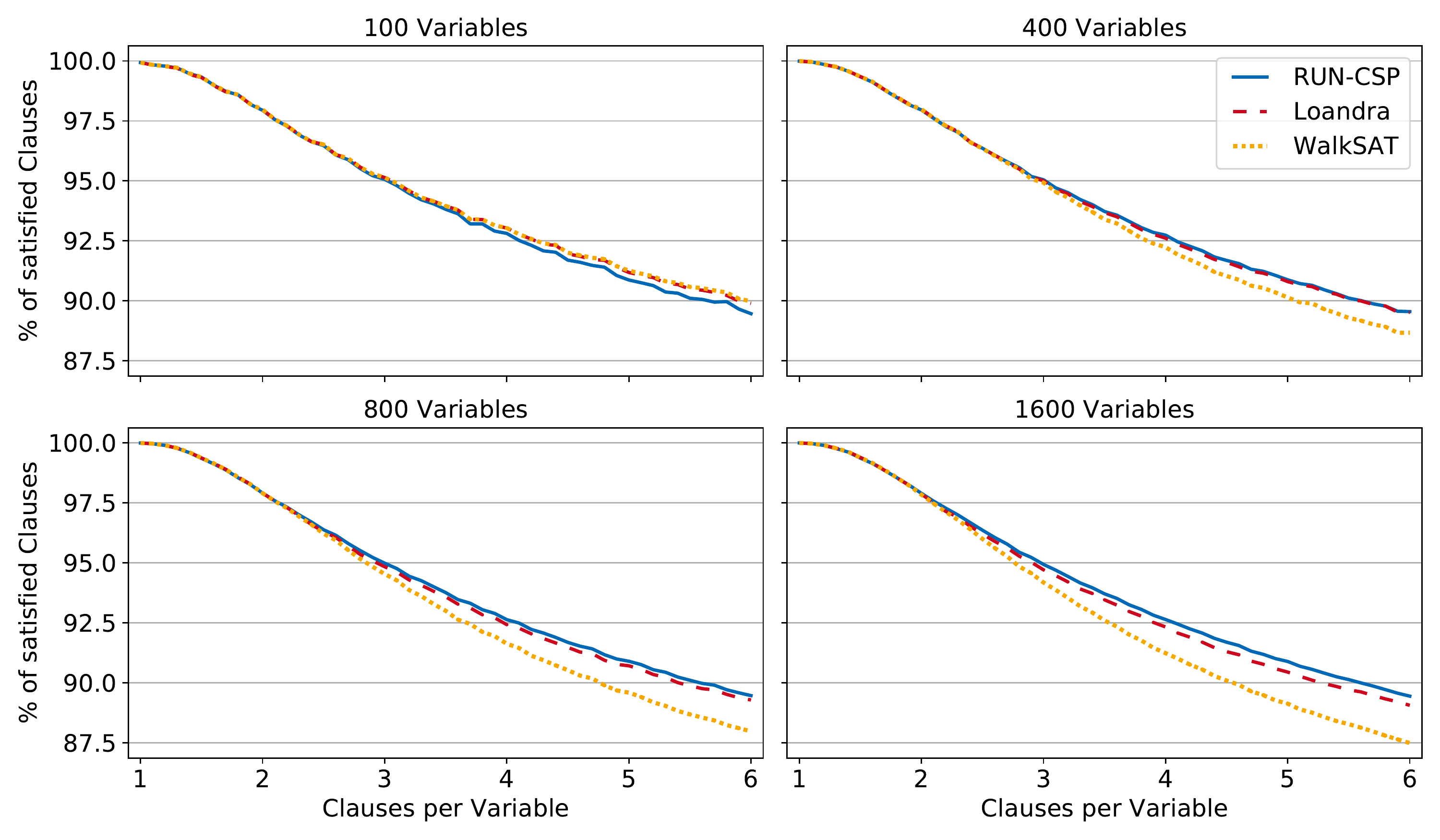}
  \end{center}
  \caption{Percentage of satisfied clauses of random \textsc{2-CNF} formulas for \network{}, Loandra and WalkSAT.
  Each data point is the average of \(100\) formulas; the ratio of clauses per variable increases in steps of \(0.1\).}
  \label{fig:2sat-opt}
\end{figure*}

We ran our experiments on machines with two Intel Xeon 8160 CPUs and one NVIDIA Tesla V100 GPU but got very similar runtimes on consumer hardware.
Evaluating 64 runs on an instance with 1,000 variables and 1,000 constraints takes about 1.5 seconds, 10,000 constraints about 5 seconds, and 20,000 constraints about 8 seconds.
Training a model takes less than $30$ minutes.
Thus, the computational cost of \network{} is relatively low.

\subsection{Max-2-SAT}

We view \msat{} as a binary CSP with domain \mbox{$D\!=\!\{0,1\}$} and a constraint language consisting of three relations $R_{00}$ (for clauses with two negated literals), $R_{01}$ (one negated literal) and $R_{11}$ (no negated literals). 
For example, $R_{01}=\{(0,0),(0,1),(1,1)\}$ is the set of satisfying assignments for a clause $(\neg x\vee y)$.
For training a \network{} model we used 4,000 random \textsc{2-CNF} formulas with $100$ variables each.
The number of clauses was sampled uniformly between $100$ and $600$ for every formula and each clause was generated by sampling two distinct variables and then independently negating the literals with probability $0.5$.

For the evaluation of \network{} in \msat{} we start with random instances and compare it to a number of problem-specific heuristics.
All baselines can solve \textsc{Max-Sat} for arbitrary arities, not only \msat{}, while \network{} can solve a variety of binary \mcsp{}s.
The state-of-the-art \textsc{Max-Sat} Solver \emph{Loandra} \cite{berg2019core} won the unweighted track for incomplete solvers in the \textsc{Max-Sat} Evaluation 2019 \cite{bacchus2019maxsat}.
We ran Loandra in its default configuration with a timeout of 20 minutes on each formula.
To put this into context, on the largest evaluation instance used here (9,600 constraints) \network{} takes less than seven minutes on a single CPU core and about five seconds using the GPU.
\emph{WalkSAT} \cite{selman1993local,WalkSat} is a stochastic local search algorithm for approximating \textsc{Max-Sat}.
We allowed WalkSAT to perform 10 million flips on each formula using its `noise' strategy with parameters $n=2$ and $m=2000$.
Its performance was boosted similarly to \network{} by performing 64 runs and selecting the best result.

For evaluation we generated random formulas with 100, 400, 800 and 1,600 variables.
The ratio between clauses and variables was varied in steps of $0.1$ from $1$ to $6$.
Figure \ref{fig:2sat-opt} shows the average percentage of satisfied clauses in the solutions found by each method over $100$ formulas for each size and density.
The methods yield virtually identical results for formulas with less than 2 clauses per variable.
For denser instances, \network{} yields slightly worse results than both baselines when only 100 variables are present.
However, it matches the results of Loandra for formulas with 400 variables and outperforms the solver for instances with 800 and 1,600 variables.
The performance of WalkSAT degrades on these formulas and is significantly worse than \network{}.

\begin{table}[t]
	\centering
	\caption{Number of unsatisfied constraints for \msat{} benchmark instances derived from the Ising spin glass problem. Best solutions are printed in bold.}
	\label{tab:2SAT_Benchmarks}
	\begin{tabular}{@{}lcccccc@{}}
		\toprule			
	Instance\!\!\!\!\! & $|V|$ & $|C|$ & \!Opt.\! & \!\!\!\!\network{}\!\!\!& \!WalkSAT\!\! & \!\!\!Loandra\\
		\midrule
		 t3pm3\!\!\! & 27 & \!162\! & 17 & \textbf{17} & \textbf{17} & \!\!\!\textbf{17} \\	
		 t4pm3\!\!\! & 64 & \!384\! & 38 & 40 & \textbf{38} & \!\!\!\textbf{38} \\	
		 t5pm3\!\!\! & 125 & \!750 & 78 & \textbf{78} & \textbf{78} & \!\!\!\textbf{78} \\	
		 t6pm3\!\!\! & 216 & \!\!1269\!\! & \!\!136\!\! & \textbf{136} & 142 & \!\!\!142 \\
		 t7pm3\!\!\! & 343 & \!\!2058\!\! & \!\!209\!\! & \textbf{216} & 227 & \!\!\!225 \\
		\bottomrule
 	\end{tabular}
\end{table}
 
For more structured formulas, we use \msat{} benchmark instances from the unweighted track of the \textsc{Max-Sat} Evaluation 2016 \cite{MaxSat2016} based on the Ising spin glass problem \cite{de1995exact,heras20082006}.
We used the same general setup as in the previous experiment but increased the timeout for Loandra to $60$ minutes.
In particular we use the same \network{} model trained entirely on random formulas.
Table \ref{tab:2SAT_Benchmarks} contains the achieved numbers of unsatisfied constraints across the benchmark instances.
All methods produced optimal results on the first and the third instance.
\network{} slightly deviates from the optimum on the second instance.
For the fourth instance \network{} found an optimal solution while both WalkSAT and Loandra did not.
On the largest benchmark formula, \network{} again produced the best result.

Thus, \network{} is competitive for random as well as spin-glass-based structured \msat{} instances.
Especially on larger instances it also outperforms conventional methods.
Furthermore, training on random instances generalized well to the structured spin-glass instances.

\subsection{Max Cut}
\begin{table}[t]
	\centering
	 \caption{$P$-values of graph cuts produced by \network{}, Yao, SDP, and EO for regular graphs with $500$ nodes and varying degrees. We report the mean across $1000$ random graphs for each degree.
    }
	\label{tab:Max-Cut}
	\begin{tabular}{@{}cccccc@{}}
		\toprule			
    d & \network{} & Yao Rel. & Yao Pol. & SDP & EO\\
		\midrule
		 3 & 0.714 & 0.707 & 0.693 & 0.702 & 0.727 \\	
		 5 & 0.726 & 0.701 & 0.668 & 0.690 & 0.737 \\	
		 10 & 0.710 & 0.670 & 0.599 & 0.682 & 0.735 \\	
		 15 & 0.697 & 0.607 & 0.629 & 0.678 & 0.736 \\	
		 20 & 0.685 & 0.614 & 0.626 & 0.674 & 0.732 \\
		\bottomrule
 	\end{tabular}
\end{table}
 
\mcut{} is a classical \mcsp{} with only one relation $\{(0,1),(1,0)\}$ used in the constraints.
In this section we evaluate \network{}'s performance on this problem.
\citet{yao2019experimental} proposed two unsupervised \gnn{} architectures for \mcut{}.
One was trained through policy gradient descent on a non-differentiable loss function while the other used a differentiable relaxation of this loss.
They evaluated their architectures on random regular graphs, where the asymptotic \mcut{} optimum is known.
We use their results as well as their baseline results for Extremal Optimization (EO) \cite{boettcher2001extremal} and a classical approach based on semi-definite programming (SDP) \cite{goemans1995improved} as baselines for \network{}. 
To evaluate the sizes of graph cuts \citeinline{yao2019experimental} introduced a relative performance measure called \textit{P-value} given by $P(z) = \frac{z/n - d/4}{\sqrt{d/4}}$ where $z$ is the predicted cut size for a $d$-regular graph with $n$ nodes.
Based on results of \citeinline{dembo2017extremal}, they showed that the expected $P$-value of $d$-regular graphs approaches $P^* \approx 0.7632$ as $n \rightarrow \infty$. 
$P$-values close to $P^*$ indicate a cut where the size is close to the expected optimum and larger values are better.
While \citeauthor{yao2019experimental} trained one instance of their GNN for each tested degree, we trained one network model on 4,000 \er{} graphs and applied it to all graphs.
For training, each graph had a node count of $n=100$ and a uniformly sampled number of edges $m \sim U(100,2000)$.
Thus, the model was not trained specifically for regular graphs.
Table \ref{tab:Max-Cut} reports the mean $P$-values across 1,000 random regular graphs with 500 nodes for different degrees. For every method other than \network{}, we provide the values as reported by \citeauthor{yao2019experimental}.
While \network{} does not match the cut sizes produced by extremal optimization, it clearly outperforms both versions of the \gnn{} as well as the classical SDP-based approach.

We performed additional experiments on standard \mcut{} benchmark instances.
The Gset dataset \cite{Gset} is a set of 71 weighted and unweighted graphs that are commonly used for testing \mcut{} algorithms.
The dataset contains three different types of random graphs.
Those graphs are \er{} graphs with uniform edge probability, graphs where the connectivity gradually decays from node $1$ to $n$, and $4$-regular toroidal graphs. Here, we use two unweighted graphs for each type from this dataset.

\begin{table}[t]
	\centering
	\caption{Achieved cut sizes on Gset instances for \network{}, DSDP and BLS.}
	\label{tab:MaxCut_Gset}
	\begin{tabular}{@{}lccccc@{}}
		\toprule			
	Graph & $|V|$ & $|E|$ & \network{} & DSDP & BLS \\
		\midrule
		 G14 & 800 & 4694 & 2943 & 2922 &  \textbf{3064} \\
		 G15 & 800 & 4661 & 2928 & 2938 &  \textbf{3050} \\
		 G22 & 2000 & 19990 & 13028 & 12960 &  \textbf{13359} \\
		 G49 & 3000 & 6000 & \textbf{6000} & \textbf{6000} & \textbf{6000} \\
		 G50 & 3000 & 6000 & \textbf{5880} & \textbf{5880} & \textbf{5880} \\
		 G55 & 5000 & 12468 & 10116 & 9960 & \textbf{10294} \\	
		 \bottomrule
 	\end{tabular}
\end{table}
 
We reused the \network{} model from the previous experiment but increased the number of iterations for evaluation to $\eval{}=500$.
Our first baseline by \citeinline{choi2000solving} uses an SDP solver based on dual scaling (DSDP) and a reduction based on the approach of \citeinline{goemans1995improved}.
Our second baseline Breakout Local Search (BLS) is based on the combination of local search and adaptive perturbation \cite{BENLIC20131162}.
Its results are among the best known solutions for the Gset dataset.
For DSDP and BLS we report the values as provided in the literature.
Table \ref{tab:MaxCut_Gset} reports the achieved cut sizes for \network{}, DSDP, and BLS.
On G14 and G15, which are random graphs with decaying node degree, the graph cuts produced by \network{} are similar in size to those reported for DSDP.
For the \er{} graphs G22 and G55 \network{} performs better than DSDP but worse than BLS.
Lastly, on the toroidal graphs G49 and G50 all three methods achieved the best known cut size.
This reaffirms the observation that our architecture works particularly well for regular graphs.
Although \network{} did not outperform the state-of-the-art heuristic in this experiment it performed at least as well as the SDP based approach DSDP.

\subsection{Coloring}
Within coloring we focus on the case of $3$ colors, \ie{} we consider CSPs over the domain $\{1,2,3\}$ with one constraint relation $\{(i,j);i,j \in D, i \neq j\}$.
In general, \network{} aims to satisfy as many constraints as possible and therefore approximates \mcol{}.
Instead of evaluating on \mcol{}, we evaluate on its practically more relevant decision variant \dcol{} which asks whether a given graph is 3-colorable without conflicts.
We turn \network{} into a classifier by predicting that a given input graph is 3-colorable if and only if it is able to find a conflict-free vertex coloring.

\begin{table}[t]
	\centering
	 \caption{
      Percentages of hard instances classified correctly by \network{}, Greedy, HybridEA, and GNN-GCP.
We evaluate on 1,000 instances for each size.
      We provide mean and standard deviation across five different \network{} models.
    }
	\label{tab:3-COL-Solved}
	\begin{tabular}{@{}cccc|cc@{}}
		 \toprule			

		 Nodes & \network{} & Greedy\! & \!HybridEA & \multicolumn{2}{c}{GNN-GCP} \\
		 & Pos. & Pos. & Pos. & Pos. & Neg.\\
		 \midrule
		 50 & 98.4\,$\pm$0.3 & 34.0 & 100.0 & 77.6 & 27.0 \\
		 100 & 62.5\,$\pm$2.7 & 6.7 & 100.0 & 64.5 & 37.8 \\
		 150 & 15.5\,$\pm$2.3 & 1.5 & 98.7 & 57.7 & 43.5 \\
		 200 & 2.6\,$\pm$0.4 & 0.5 & 88.9 & 51.9 & 45.3 \\
		 300 & 0.1\,$\pm$0.0 & 0.0 & 39.9 & 48.8 & 52.7 \\
		 400 & 0.0\,$\pm$0.0 & 0.0 & 15.3 & 46.3 & 54.7 \\		
		\bottomrule
 	\end{tabular}
\end{table}
 
We evaluate \network{} on so-called `hard' random instances, similar to those defined by \citeinline{lemos2019graph}.
These instances are a special subclass of \er{} graphs where an additional edge can make the graph no longer $3$-colorable.
We describe our exact generation procedure in the appendix.
We trained five \network{} models on 4,000 hard 3-colorable instances with $100$ nodes each.
In Table \ref{tab:3-COL-Solved} we present results for \network{}, a greedy heuristic with DSatur strategy \cite{brelaz1979new}, the state-of-the-art heuristic HybridEA \cite{galinier1999hybrid, lewis2012wide, lewis2015guide}, and GNN-GCP \cite{lemos2019graph}.
We trained a GNN-GCP network on the training instances of \citeinline{lemos2019graph} and allowed HybridEA to make 500 million constraint checks on each graph.
All algorithms but GNN-GCP report a graph a 3-colorable only if they found a valid 3-coloring on it and thus never produce false positives.
Thus, we additionally report values for GNN-GCP on the non-3-colorable counterparts of the evaluation instances.
For GNN-GCP we observe that it roughly gets the chromatic number right and only predicts 3 or 4 on our test graphs.
Nevertheless, it seems unable to correctly classify the test instances as high accuracy on positive instances comes with low accuracy on negative instances and vice versa.
For the other three algorithms we observe a clear hierarchy.
HybridEA expectedly performs best and finds solutions even for some of the largest graphs.
\network{} correctly classifies most of the graphs up to 100 nodes and clearly outperforms GNN-GCP.
The weakest algorithm is DSatur which even fails on most of the small 50 node graphs.

Using a \network{} model trained on a mixture of random graphs performs slightly worse than models trained on hard random graphs only (shown in the appendix).

Overall, we see that despite being designed for maximization tasks, \network{} outperforms greedy heuristics and neural baselines on the decision variant of \dcol{} for hard random instances.

\subsection{Independent Set}

\begin{figure*}[t]
\includegraphics[width=1.0\textwidth]{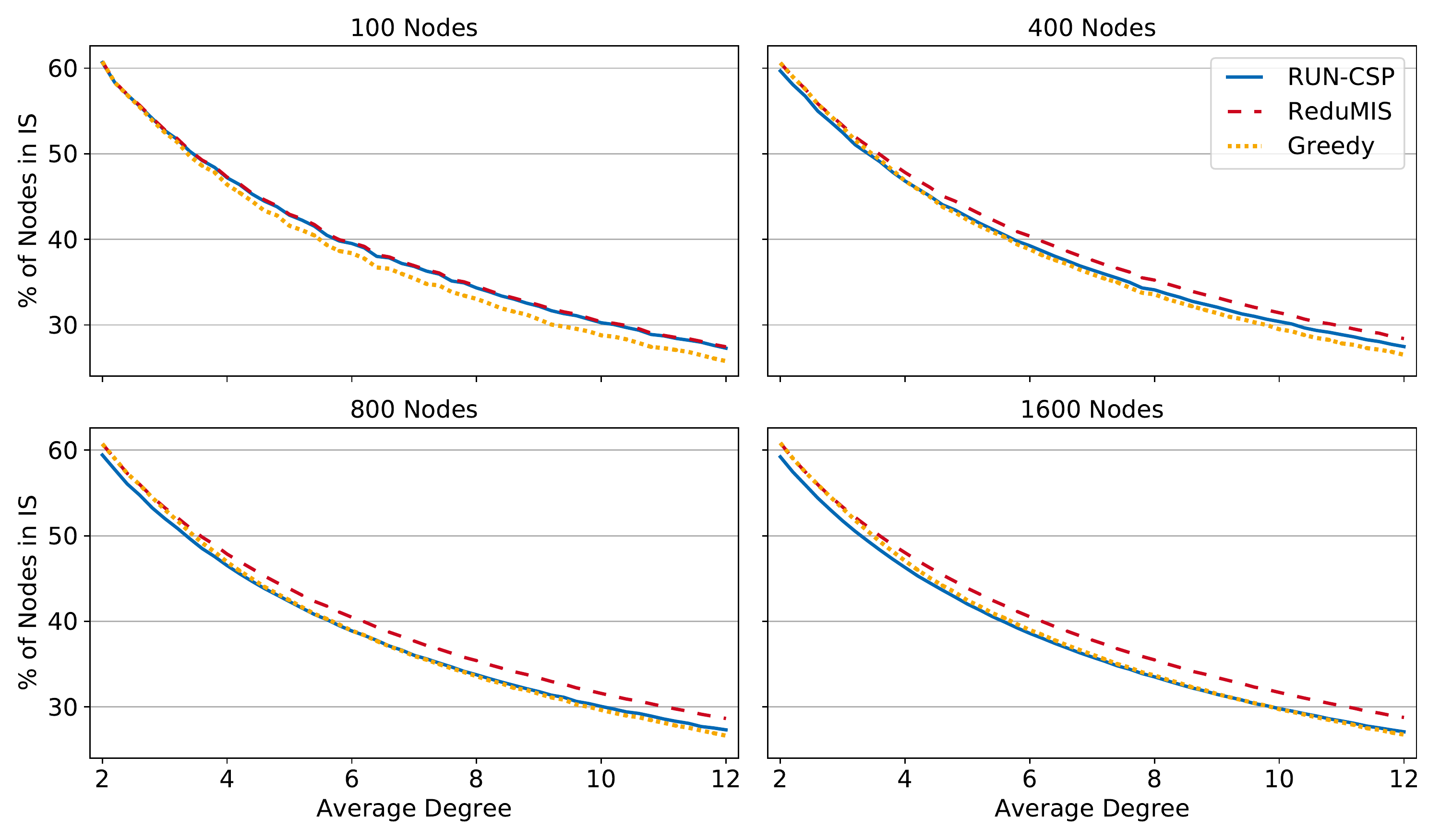}
\caption{Independent set sizes on random graphs produced by \network{}, ReduMIS and a greedy heuristic.
	The sizes are given as the percentage of nodes contained in the independent set.
	Every data point is the average for 100 graphs; the degree increases in steps of \(0.2\).}
	\label{fig:abs_is_opt}
\end{figure*}
\enlargethispage{0.5cm}
Finally, we experimented with the maximum independent set problem \mis{}.
The independence condition can be modeled through a constraint language $\Gamma_{\text{IS}}$ with one binary relation $R_{\text{IS}}=\{(0,0), (0,1), (1,0)\}$.
Here, assigning the value $1$ to a variable is equivalent to including the corresponding node in the independent set.
\mis{} is \emph{not} simply \mcsp{}($\Gamma_{\text{IS}}$), since the empty set will trivially satisfy all constraints.
Instead, \mis{} is equivalent to finding an assignment which satisfies $R_{\text{IS}}$ at all edges while maximizing an additional objective function that measures the size of the independent set.
To model this in our framework, we extend the loss function to reward assignments with many variables set to 1.
For a graph $G=(V,E)$ and a soft assignment $\phi:V\to[0,1]$, we define
\begin{align}
	\mathcal{L}_\text{MIS}(\varphi, G) \!&=\big(\kappa\! + \!\mathcal{L}_\text{CSP}(\varphi, G)\big) \cdot 
	\big(1\!+\!\mathcal{L}_\text{size}(\varphi,G) \big),\label{eq:is}\\
\mathcal{L}_\text{size}(\varphi, G\big)&=\frac{1}{|V|}\sum_{v\in V}(1 - \phi(v)).\nonumber
\end{align}
Here, $\mathcal{L}_\text{CSP}$ is the standard \network{} loss for $\Gamma_\text{IS}$ and
$\kappa$ adjusts the relative importance of $ \mathcal{L}_\text{CSP}$ and $ \mathcal{L}_\text{size}$.
Intuitively, smaller values for $\kappa$ decrease the importance of $\mathcal{L}_\text{size}$ which favors larger independent sets.
A naive weighted sum of both terms turned out to be unstable during training and yielded poor results, whereas the product in Equation \eqref{eq:is} worked well.
For training, $\mathcal{L}_\text{MIS}$ is combined across iterations with a discount factor $\lambda$ as in the standard \network{} architecture.

We start by evaluating the performance on random graphs.
We trained a network on 4,000 random \er{} graphs with $100$ nodes and $m \sim U(100, 600)$ edges each and with $\kappa=1$.
For evaluation we use random graphs with $100$, $400$, $800$ and 1,600 nodes and a varying number of edges.
For roughly $6\%$ of all predictions, the predicted set contained induced edges (just a single edge in most cases), meaning the predicted sets where not independent.
We corrected these predictions by removing one of the endpoints of each induced edge from the set and only report results after this correction.
We compare \network{} against two baselines: ReduMIS, a state-of-the-art \mis{} solver \cite{lamm2017finding, akiba2016branch} and a greedy heuristic, which we implemented ourselves.
The greedy procedure iteratively adds the node with lowest degree to the set and removes the node and its neighbors from the graph until the graph is empty.
Figure \ref{fig:abs_is_opt} shows the achieved independent set sizes, each data point is the mean IS size across $100$ random graphs.
For graphs with $100$ nodes, \network{} achieves similar sizes as ReduMIS and clearly outperforms the greedy heuristic.
On larger graphs our network produces smaller sets than ReduMIS. 
However, \network{}'s performance remains similar to the greedy baseline and, especially on denser graphs, outperforms it.

For more structured instances, we use a set of benchmark graphs from a collection of hard instances for combinatorial problems \cite{Xu2005Benchmarks}. The instances are divided into five sets with five graphs each.
These graphs were generated through the RB Model \cite{xu2003many, xu2005simple}, a model for generating hard CSP instances. A graph of the class frb$c$-$k$ consists of $c$ interconnected $k$-cliques and
the maximum independent set has a forced size of $c$.
The previous model trained on \er{} graphs did not perform well on these instances and produced sets with many induced edges.
Thus, we trained a new network on 2,000 instances we generated ourselves through the RB Model.
The exact generation procedure of this dataset is provided in the appendix.
Training used a batch size of 5 for 25 epochs.
We set $\kappa=0.1$ to increase the importance of the independence condition.
The predictions of the new model contained no induced edges for all benchmark instances.
Table \ref{tab:IS_Benchmarks} contains the achieved IS sizes.
\network{} yields similar results as the greedy heuristic.
While our network does not match the state-of-the-art heuristic, it still beats the greedy approach on instances with over 100,000 edges.
\begin{table}[t]
	\centering
	 \caption{Achieved \textsc{IS} sizes for the benchmark graphs. We report the mean and std.\ deviation for the 5 graphs in each group.}
	\label{tab:IS_Benchmarks}
	\begin{tabular}{@{}lccccc@{}}
		\toprule			
	   Graphs\!\! & \!\!$|V|$\!\!\! & \!\!$|E|$\!\! & \!\!\!\network{}\!\!& \!\!Greedy\!\! & \!\!ReduMIS\\
		\midrule
		frb30-15\!\! & \!\!450\!\!\! & \!\!18k\!\! & 25.8\,$\pm$0.8 & 24.6\,$\pm$0.5 & 30\,$\pm$0.0 \\	
		frb40-19\!\! & \!\!790\!\!\! & \!\!41k\!\! & 33.6\,$\pm$0.5 & 33.0\,$\pm$1.2 & 39.4\,$\pm$0.5 \\	
		frb50-23\!\! & \!\!1150\!\!\! & \!\!80k\!\! & 42.2\,$\pm$0.4 & 42.2\,$\pm$0.8 & 48.8\,$\pm$0.4 \\	
		frb59-26\!\! & \!\!1478\!\!\! & \!\!126k\!\! & 49.4\,$\pm$0.5 & 48.0\,$\pm$0.7 & 57.4\,$\pm$0.9 \\			
		\bottomrule
 	\end{tabular}
\end{table}

\section{Conclusions}
We have presented a universal approach for approximating \mcsp{}s with recurrent neural networks.
Its key feature is the ability to train without supervision on any available data.
Our experiments on the optimization problems \msat{}, \mcut{}, \dcol{} and \mis{} show that \network{} produces high quality approximations for all four problems.
Our network can compete with traditional approaches like greedy heuristics or semi-definite programming on random data as well as benchmark instances.
For \msat{}, \network{} was able to outperform a state-of-the-art \textsc{Max-Sat} Solver.
Our approach also achieved better results than neural baselines, where those were available.
\network{} networks trained on small random instances generalize well to other instances with larger size and different structure.
Our approach is very efficient and inference takes a few seconds, even for larger instances with over 10,000 constraints.
The runtime scales linearly in the number of constraints and our approach can fully utilize modern hardware, like GPUs.

Overall, \network{} seems like a promising approach for approximating \mcsp{}s with neural networks.
The strong results are somewhat surprising, considering that our networks consist of just one LSTM Cell and a few linear functions.
We believe that our observations point towards a great potential of machine learning in combinatorial optimization.

\paragraph{Future Work}
We plan to extend \network{} to CSPs of arbitrary arity and to weighted \csp{}s. 
It will be interesting to see, for example, how it performs on 3-\textsc{SAT} and its maximization variant.
Another possible future extension could combine \network{} with traditional local search methods, similar to the approach by \citeinline{li2018combinatorial} for \mis{}.
The soft assignments can be used to guide a tree search and the randomness can be exploited to generate a large pool of initial solutions for traditional refinement methods.

\pagebreak

\printbibliography 

\pagebreak
\appendix

\section{External Software and Data}
Table \ref{tab:Software} lists the versions and sources of all the external software used in our paper.
Table \ref{tab:Data} lists the sources of all the evaluation instances that we did not generate ourselves.
All evaluation instances are also provided in our repository at \url{https://github.com/RUNCSP/RUN-CSP}
\begin{table}[H]
	\centering
	\caption{Used Software}
	\label{tab:Software}
	\begin{tabular}{@{}lll@{}}
		\toprule			
		Software & Version & Author / Source \\
		\midrule
		Tensorflow & 1.15.2 & \cite{tensorflow2015-whitepaper} \\
		& & \small\url{https://tensorflow.org} \\
		\midrule
		NetworkX & 2.4 & \cite{team2014networkx} \\
		& & \small\url{https://networkx.github.io} \\
		\midrule
		Loandra & 2019 & \cite{berg2019core} \\
		& & \small\url{https://maxsat-evaluations.github.io/2019/descriptions} \\
		\midrule
		MaxWalkSAT & 20 & \cite{selman1993local} \\
		& & \small\url{https://www.cs.rochester.edu/u/kautz/walksat/}\\
		\midrule
		ReduMIS & 1.1 & \cite{kamis2019JV} \\
		& & \small\url{http://algo2.iti.kit.edu/kamis/} \\
		\midrule
		HybridEA & 2015 & \cite{galinier1999hybrid}, \cite{lewis2012wide}, \cite{lewis2015guide} \\
		& & \small\url{http://rhydlewis.eu/resources/gCol.zip} \\
		\midrule
		GNN-GCP & 2020 & \cite{lemos2019graph} \\
		& & \small\url{https://github.com/machine-reasoning-ufrgs/GNN-GCP} \\
		\bottomrule
 	\end{tabular}
\end{table}
 \begin{table}[H]
	\centering
	\caption{Used Data}
	\label{tab:Data}
	\begin{tabular}{@{}ll@{}}
		\toprule			
		Instances & Author / Source \\
		\midrule
		Spinglass & \cite{heras20082006} \\
		\textsc{2-CNF} & \small\url{http://maxsat.ia.udl.cat/benchmarks/} , \\
		& Unweighted Crafted Benchmarks \\
		\midrule
		Gset & \cite{Gset} \\
		& \small\url{https://www.cise.ufl.edu/research/sparse/matrices/Gset/} \\
		\midrule
		\mis{} & \cite{Xu2005Benchmarks} \\ 
		Graphs & \small\url{http://sites.nlsde.buaa.edu.cn/~kexu/benchmarks/graph-benchmarks.htm} \\
		\bottomrule
 	\end{tabular}
\end{table}

\section{Reproducibility}
This section summarizes the necessary information for replicating our implementation and exact training procedure.
The information can also be found in training scripts provided in the source code.
All trainable matrices in the messaging functions $S_R$ were initialized with uniform Glorot initialization.
For activation, bias, and initialization of the LSTM cell we used the default values provided by TensorFlow 1.14.0.
All trainable parameters were regularized with the $\ell^2$-norm with a weight of $0.01$.
The pooled vectors of received messages $r_x^{(t)}$ were normalized with an additional batch normalization layer, before being passed into the LSTM cell.
The discount factor in our loss function was set to $\lambda=0.95$ in all experiments.
The state size of all networks was set to $k=128$.

All networks were trained with $\train{}=30$.
Training was performed with the Adam optimizer using the default parameters $\beta_1=0.9$, $\beta_2=0.999$, and $\epsilon=1 \times 10^{-7}$.
The learning rate was initialized as $0.001$ and decayed with a factor of $0.1$ every $5$ epochs.
The gradients were clipped at a norm of $1.0$.

\subsection{Parallel execution}

We perform 64 runs on each instance to boost the performance during evaluation.
Our implementation performs these runs in parallel with a single forward pass of the network.
To achieve this, we copy the graph 64 times and combine the disjoint copies into one larger instance.
This instance is then processed by the \network{} network.
Since there are no messages exchanged between connected components, this is equivalent to executing the network 64 times on the same graph. 
Evaluating on a single large graph is, especially for small instances, faster on a GPU than multiple executions on the small graphs.

For the largest IS benchmark graphs the parallel inference with 64 copies exceeds the memory limit of our GPU.
Instead, we performed 8 consecutive runs with 8 copies each and selected the best assignment afterwards.

\section{Hard 3-Col Instances}
Here, we describe how the `hard' 3-col graphs used in our vertex coloring experiment are generated.
These instances are 3-colorable \er{} graphs and there is (at least) one edge $e$ such that when adding $e$ to the graph, it is non-3-colorable.
To generate such a graph, we first initialize a graph with $n$ nodes and no edges.
We then iteratively add individual edges which are uniformly sampled at random.
After adding each edge, we use a conventional \sat{} solver to check whether the graph is still 3-colorable. (We use the pycosat Python package for this task.)
This process is stopped once the graph is found to be non-3-colorable.
The graphs with and without the last added edge are then returned as negative and positive instances, respectively.
Our graph generation procedure stops adding edges when the next randomly chosen edge makes the graph non-3-colorable.
In contrast, \citet{lemos2019graph} stopped adding edges when it was possible to add such an edge that makes the graph non-3-colorable.
The graphs generated by \citeauthor{lemos2019graph} are thus less dense than ours.

\section{\mis{} Benchmarks}
The \network{} network that was evaluated on the \mis{} benchmark graphs was trained on our own synthetic benchmark instances.
Here, we will describe the generation procedure of these instances.
\cite{xu2003many, xu2005simple} proposed the RB Model, which is a general model for generating hard random CSP instances close to the phase change of satisfiability.
Furthermore, they described how to generate hard instances for graph problems, including \mis{} instances, by reducing SAT benchmarks of the RB Model to these problems.
We used their generation procedure for \mis{} benchmarks as described in \citeinline{Xu2005Benchmarks}.
Given $c \in \NN$, $p \in [0,1]$ and $\alpha, r > 0$, the procedure generates a graph as follows:
\begin{enumerate}
	\item Generate $c$ disjoint cliques with $k=c^{\alpha}$ vertices each.
	\item Select two random cliques and generate $pc^{2\alpha}$ random edges between them (without repetition).
	\item Run Step 2 for another $rc \ln c - 1$ times (with repetition).
\end{enumerate}
To enforce an optimal independent set of size $c$, one can exclude one node of each clique from the process of adding random edges.

We used this procedure to generate 2,000 training instances.
For each graph we uniformly sampled $c \sim U(10, 25)$, $k \sim U(5, 20)$ and $p \sim U(0.3, 1.0)$.
We then chose
\begin{equation}
\alpha = \frac{\ln(k)}{\ln(c)} \quad\text{and}\quad r = - \frac{\alpha}{\ln(1 - p)}
\end{equation}
This choice for $r$ is expected to yield `hard' instances according to the RB Model \cite{xu2005simple}.
We then used the algorithm described above to generate a graph with the chosen parameters.

Training was performed with a batch size of $5$.
Note that we reduced the batch size in comparison to all other experiments due to the relatively large size of the graphs.
The constant $\kappa$ that distributes the importance of the losses $\mathcal{L}_{\text{CSP}}$ and $\mathcal{L}_{\text{size}}$ was reduced to $\kappa = 0.1$ to emphasize the independence condition. 
Without this reduction the computed solutions contained multiple edges violating independence and not just one or two as in the other IS experiment.

\section{Additional Experiments}

\subsection{Detailed Gset Results}
Table \ref{tab:MaxCut_Gset} provides the achieved cut sizes for additional Gset instances.
As for the subset of instances provided in the Experiments Section, the \network{} network performed $64$ parallel evaluation runs with $\eval{}=500$ iterations.

\begin{table}[t]
	\centering
	 \caption{Achieved cut sizes on Gset instances for \network{}, DSDP and BLS. The values for DSDP and BLS taken from the literature. DSDP values were only available for a subset of GSET.}
	\label{tab:MaxCut_Gset}
	\begin{tabular}{@{}lccc|ccc@{}}
		\toprule			
	Graph & type & $|V|$ & $|E|$ & \network{} & DSDP & BLS \\
		\midrule
		 G1 & \textit{random} & 800 & 19176 & 11369 & - & \textbf{11624} \\
		 G2 & \textit{random} & 800 & 19176 & 11367 & - & \textbf{11620} \\
		 G3 & \textit{random} & 800 & 19176 & 11390 & - & \textbf{11622} \\
		 G14 & \textit{decay} & 800 & 4694 & 2943 & 2922 & \textbf{3064} \\
		 G15 & \textit{decay} & 800 & 4661 & 2928 & 2938 & \textbf{3050} \\
		 G16 & \textit{decay} & 800 & 4672 & 2921 & - & \textbf{3052} \\
		 G22 & \textit{random} & 2000 & 19990 & 13028 & 12960 & \textbf{13359} \\
		 G23 & \textit{random} & 2000 & 19990 & 13006 & 13006 & \textbf{13344} \\
		 G24 & \textit{random} & 2000 & 19990 & 13001 & 12933 & \textbf{13337} \\
		 G35 & \textit{decay} & 2000 & 11778 & 7339 & - & \textbf{7684} \\
		 G36 & \textit{decay} &  2000 & 11766 & 7325 & - & \textbf{7678} \\
		 G37 & \textit{decay} &  2000 & 11785 & 7317 & - & \textbf{7689} \\
		 G48 & \textit{torus} & 3000 & 6000 & \textbf{6000} & \textbf{6000} & \textbf{6000} \\
		 G49 & \textit{torus} & 3000 & 6000 & \textbf{6000} & \textbf{6000} & \textbf{6000} \\
		 G50 & \textit{torus} & 3000 & 6000 & \textbf{5880} & \textbf{5880} & \textbf{5880} \\
		 G51 & \textit{decay} & 1000 & 5909 & 3690 & - & \textbf{3848} \\
		 G52 & \textit{decay} & 1000 & 5916 & 3681 & - & \textbf{3851} \\
		 G53 & \textit{decay} & 1000 & 5914 & 3695 & - & \textbf{3850} \\
		 G55 & \textit{random} & 5000 & 12468 & 10116 & 9960 & \textbf{10294} \\	
		 \bottomrule
 	\end{tabular}
\end{table}

\subsection{Structure Specific Performance for Max-3-Col}
\enlargethispage{0.5cm}
A key feature of \network{} is the unsupervised training procedure, which allows us to train a network on arbitrary data without knowledge of any optimal solutions.
Intuitively, we would expect a network trained on instances of a particular structure to adapt toward this class of instances and perform poorer for different structures.
We will briefly evaluate this hypothesis for four different classes of graphs using the \mcol{} problem.
\begin{description}
	\item[\er{} Graphs:] Graphs are generated by uniformly sampling $m$ distinct edges between $n$ nodes.
	\item[Geometric Graphs:] A graph is generated by first assigning random positions within a $1 \times 1$ square to $n$ distinct nodes. Then an edge is added for every pair of points with a distance less than $r$.
	\item[Powerlaw-Cluster Graphs:] This graph model was introduced by \citeinline{holme2002growing}. 
Each graph is generated by iteratively adding $n$ nodes. 
	Every new node is connected to $m$ random nodes that were already inserted.
	After each edge is added, a triangle is closed with probability $p$, \ie{} an additional edge is added between the inserted node and a random neighbor of the other endpoint of the edge.
	\item[Regular Graphs:] We consider random 5-regular graphs as an example for graphs with very specific structure class. 
\end{description}
For each graph class we generated a training dataset with 4,000 random instances.
The number of nodes was sampled uniformly between $50$ and $100$ for each graph of all four classes.
For \er{} graphs, the edge count $m$ was chosen randomly between $100$ and $400$.
The parameter $r$ of each geometric graph was sampled uniformly from the interval $[0.1,0.2]$.
For Powerlaw-Cluster graphs, the parameter $m$ was uniformly sampled from $\{1,2,3\}$ and $p$ was uniformly drawn from the interval $[0,1]$.
Five \network{} models were trained on each dataset.
We refer to these groups of models as $M_\text{ER}$, $M_\text{Geo}$, $M_\text{Pow}$ and $M_\text{Reg}$.
Additionally, 5 models $M_\text{Mix}$were trained on a mixed dataset with 1,000 random instances of each graph class.

For evaluation, we generated 1,000 instances of each class.
Table \ref{tab:3-COL-Structure} contains the percentage of unsatisfied constraints over the models and graph classes.
In general, all models perform well on the class of structures they were trained on.
$M_\text{Geo}$ and $M_\text{Pow}$ outperform $M_\text{ER}$ on \er{} graphs while $M_\text{ER}$ outperforms $M_\text{Geo}$ on Powerlaw-Cluster graphs and $M_\text{Pow}$ on geometric graphs.
The networks trained on $5$-regular graphs only perform well on the same class and yield poor results for other structures.
Overall, $M_\text{Mix}$ produced the best results when averaged over all four classes, despite not achieving the best results for any particular class.

\begin{table}[b]
	\centering
		\caption{Percentages of unsatisfied constraints for each graph class under the different \network{} models.
		Values are averaged over 1,000 graphs and the standard deviation is computed with respect to the five \network{} models.}
\label{tab:3-COL-Structure}
	\begin{tabular}{@{}l|rrrrr@{}}
		\toprule			
		Graphs & 
		\multicolumn{1}{c}{$M_\text{ER}$} & 
		\multicolumn{1}{c}{$M_\text{Geo}$} & 
		\multicolumn{1}{c}{$M_\text{Pow}$} &
		\multicolumn{1}{c}{$M_\text{Reg}$} & 
		\multicolumn{1}{c}{$M_\text{Mix}$} \\
		 & 
		\multicolumn{1}{c}{$(\%)$} & 
		\multicolumn{1}{c}{$(\%)$} & 
		\multicolumn{1}{c}{$(\%)$} & 
		\multicolumn{1}{c}{$(\%)$} & 
		\multicolumn{1}{c}{$(\%)$} \\
		\midrule
		\er{} 		& 4.75\,$\pm$0.01 & 4.73\,$\pm$0.02& \textbf{4.72}\,$\pm$0.02 & 6.69\,$\pm$1.60 & 4.73\,$\pm$0.01 \\
		Geometric 			& 10.33\,$\pm$0.07 & \textbf{10.16}\,$\pm$0.04 & 11.39\,$\pm$0.66 & 18.99\,$\pm$3.32 & 10.18\,$\pm$0.03 \\
		Pow. Cluster	& 1.89\,$\pm$0.00 & 1.96\,$\pm$0.01 & \textbf{1.87}\,$\pm$0.00 & 2.44\,$\pm$0.67 & 1.89\,$\pm$0.00 \\
		Regular 			& 2.33\,$\pm$0.01 & 2.41\,$\pm$0.03 & 2.33\,$\pm$0.02 & \textbf{2.32}\,$\pm$0.00 & 2.33\,$\pm$0.00 \\
		\midrule
		Mean 				& 4.83\,$\pm$0.02 & 4.82\,$\pm$0.03 & 5.08\,$\pm$0.18 & 7.61\,$\pm$1.40 & \textbf{4.78}\,$\pm$0.01 \\
		\bottomrule
	\end{tabular}
\end{table}

\subsection{Detailed 3-Col Results}
In Section 3.3 of the main paper we evaluated \network{} on the \dcol{} problem for hard random instances.
There, the models were trained on hard 3-colorable instances.
To further evaluate how the training data effects \network{} models, we also applied networks trained on different datasets to the same hard evaluation instances.
We trained five networks on the negative non-3-colorable counterparts of the positive training instances used earlier.
Furthermore, we applied the five models of $M_\text{Mix}$ from the previous section to the hard evaluation instances.
The achieved percentages of optimally 3-colored graphs are provided in Table \ref{tab:3-COL-Solved}.

On graphs with over 50 nodes the models trained on negative hard instance perform marginally worse than those trained on the positive counterparts.
The models of $M_\text{Mix}$ perform worse than those trained on hard instances, especially for the larger graphs.
However, they do still outperform the traditional greedy heuristic by a significant margin.

\begin{table}[t]
	\centering
	 \caption{
		Percentages of hard positive instances for which optimal 3-colorings were found by different \network{} models, Greedy Search (DSatur), and HybridEA.
		The \network{} models were trained on only positive (Pos.), only negative instances (Neg.), or  the mixed random graphs from the previous experiment.
		We report mean and standard deviation across five \network{} models that where trained on each dataset. 
	}
	\label{tab:3-COL-Solved}
	\begin{tabular}{@{}cccccc@{}}
		 \toprule			

		 Nodes & \network{} (Pos.) & \network{} (Neg.) & \network{} ($M_\text{Mix}$) & Greedy\! & \!HybridEA \\
		 \midrule
		 50 & 98.4\,$\pm$0.3 & 98.7\,$\pm$0.3 & 97.4\,$\pm$0.5 & 34.0 & 100.0 \\
		 100 & 62.5\,$\pm$2.7 & 58.9\,$\pm$3.0 & 43.2\,$\pm$2.5 & 6.7 & 100.0 \\
		 150 & 15.5\,$\pm$2.3 & 14.1\,$\pm$1.7 & 6.7\,$\pm$0.4 & 1.5 & 98.7 \\
		 200 & 2.6\,$\pm$0.4 & 1.9\,$\pm$0.3 & 0.5\,$\pm$0.3 & 0.5 & 88.9 \\
		 300 & 0.1\,$\pm$0.0 & 0.2\,$\pm$0.1 & 0.1\,$\pm$0.1 & 0.0 & 39.9 \\
		 400 & 0.0\,$\pm$0.0 & 0.0\,$\pm$0.0 & 0.0\,$\pm$0.0 & 0.0 & 15.3 \\

		\bottomrule
 	\end{tabular}
\end{table}

\subsection{Convergence of RUN-CSP}
\begin{figure}[t]
	\centering
	\includegraphics[width=1.0\textwidth]{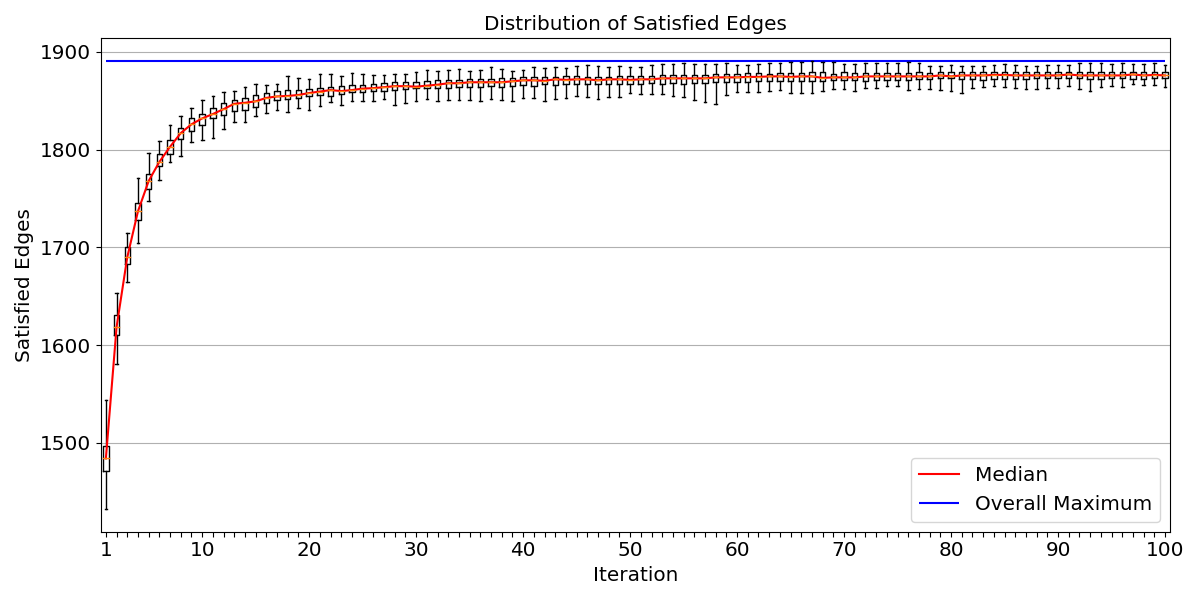}
	\caption{The distribution of satisfied edges for 64 parallel evaluation runs on a random \er{} graph. For each iteration we show a boxplot that represents the 64 numbers of satisfied edges at a given time step. The whiskers of the boxplots include the minimum and maximum values. The limits of the box are given by the upper and lower quartile. The red line represents the median in each iteration. The overall maximum (reached after 67 steps) is shown as a horizontal blue line.}
	\label{IMG:Satisfied}
\end{figure}

We now use the \mcol{} problem to illustrate the convergence behavior of \network{} networks.
We generated a random \er{} graph with 500 nodes and 2,000 edges.
One of the \network{} models of $M_\text{Mix}$ from the previous experiment was used to predict a maximum 3-coloring for this graph.
We performed 64 parallel evaluation runs for $\eval{}=100$ iterations.
For each iteration we obtain 64 values for the numbers of correctly colored edges, one for each parallel run.
Figure \ref{IMG:Satisfied} plots the distributions of these values across all iterations as boxplots.
In the first 20 time steps, the predictions improve quickly, as the network moves away from its random initialization towards a better solution.
The network continues to find better color assignments far past iteration $30$, which is the number iterations used during training, the absolute maximum was reached in iteration 67.

\subsection{Coloring Benchmark Instances}

We evaluated \network{} on a number of $k$-\textsc{COL} benchmark instances, similar to \citet{lemos2019graph}.
We obtained the 20 graphs from the COLOR02 Workshop\footnote{The graphs can be downloaded at \url{https://mat.tepper.cmu.edu/COLOR02/}} that were also used to evaluate GNN-GCP to enable a direct comparison.
Any single \network{} network is bound to a fixed domain size $d$.
A network cannot use more than $d$ colors and even if a given graph can be colored with less than $d$ colors, \network{} will still use all $d$ colors.
Thus, in order to compute a chromatic number, we trained 14 distinct networks with domain sizes ranging from 4 to 17.
We apply each network to a given graph and choose the output that achieved a conflict free coloring with the fewest colors as our final result.

Unlike our experiments for \mcut{}, we found that \network{} networks trained purely in \er{} graphs performed poorly in the given setup.
Instead, we generated mixed datasets that consist of $30\%$ \er{} graphs, $30\%$ Geometric graphs, $30\%$ Powerlaw-Cluster graphs and $10\%$ Connected Caveman graphs \cite{watts1999networks}.
Connected Caveman graphs are a graph model introduced by \citeinline{watts1999networks} that depends on two numbers $l, k \in \NN$.
All other graph classes were introduced in the Section 5.2.

The training datasets were adapted to the number $c \in \{4,\dots,17\}$ of colors available such that networks with more colors were trained on denser graphs.
Each training set contained 4,000 random graphs generated according to the following parameters:
\begin{description}
	\item[\er{}:] $n \sim U(50, 100)$, $m \sim U(2n, n \cdot c)$
	\item[Geometric:] $n \sim U(80, 28c)$, $r \sim U(0.1, 0.2)$
	\item[Powerlaw-Cluster:] $n \sim U(20, 20c)$, $m \sim U(1, 4)$, $r \sim U(1, 2)$
	\item[Connected Caveman:] $l \sim U(10, 20)$, $k \sim U(\max(4, c - 2), c + 2)$
\end{description}

We compare \network{} to the classical methods used in our previous vertex coloring experiment, namely HybridEA and the greedy DSatur strategy.
As before, HybridEA was allowed to perform 500 million constraint checks on each graph.
Table \ref{tab:benchmarkInstances3Col} provides the number of colors that each method needed to color the graphs without conflict.
For comparison, we provide the predicted chromatic number of GNN-GCP as reported by \citet{lemos2019graph}.
On most benchmark instances HybridEA finds the optimal chromatic number and otherwise uses one additional color.
For instances with a chromatic number of up to 6 the performance of our network is identical to DSatur and HybridEA.
In general, \network{} performs slightly worse than the greedy DSatur algorithm.
The number of instances for which \network{} found optimal solutions is larger than the number of graphs for which GNN-GCP predicted the correct chromatic number. 
We point out that the focus of our architecture is a maximization task associated with a domain of fixed size.
Despite this, \network{} was able to outperform GNN-GCP on this task, while also predicting color assignments.

\begin{table}
  \centering
  	\caption{Results for $k$-\textsc{Col} on benchmarks instances. We provide the number of colors needed for an optimal coloring by \network{}, Greedy (DSatur) and HybridEA. For GNN-GCP we provide the prediction as reported by \citeinline{lemos2019graph}. This method only predicts the chromatic number and can therefore underestimate the true value.}
  \label{tab:benchmarkInstances3Col}
  \begin{tabular}{@{}lcccccc@{}}
    \toprule
    Benchmark&$|V|$&Opt&\network{}&GNN-GCP&DSatur&HybridEA\\
    \midrule
    Queen5\_5&25&5&\textbf{5}&6&\textbf{5}&\textbf{5}\\
    Queen6\_6&36&7&8&\textbf{7}&8&\textbf{7}\\
    myciel5&47&6&\textbf{6}&5&\textbf{6}&\textbf{6}\\
    Queen7\_7&49&7&10&8&9&\textbf{7}\\
    Queen8\_8&64&9&11&8&10&\textbf{9}\\
    1-Insertions\_4\!\!&67&4&5&\textbf{4}&5&5\\
    huck&74&11&\textbf{11}&8&\textbf{11}&\textbf{11}\\
    jean&80&10&\textbf{10}&7&\textbf{10}&\textbf{10}\\
    Queen9\_9&81&10&17&9&12&\textbf{10}\\
    david&87&11&\textbf{11}&9&\textbf{11}&\textbf{11}\\
    Mug88\_1&88&4&\textbf{4}&3&\textbf{4}&\textbf{4}\\
    myciel6&95&7&8&\textbf{7}&\textbf{7}&\textbf{7}\\
    Queen8\_12&96&12&17&10&13&\textbf{12}\\
    games120&120&9&\textbf{9}&6&\textbf{9}&\textbf{9}\\
    Queen11\_11&121&11&$>17$&12&15&12\\
    anna&138&11&\textbf{11}&\textbf{11}&\textbf{11}&\textbf{11}\\
    2-Insertions\!\!\_4&149&4&5&\textbf{4}&5&5\\
    Queen13\_13&169&13&$>17$&14&17&14\\
    myciel7&191&8&9&NA&\textbf{8}&\textbf{8}\\
    homer&561&13&17&14&\textbf{13}&\textbf{13}\\
    \bottomrule
  \end{tabular}
\end{table}

 \end{document}